\documentclass[a4paper]{cas-sc}

\usepackage[authoryear,longnamesfirst]{natbib}
\usepackage{pgfplots}
\usepackage{multirow}
\usepackage{graphicx}
\usepackage{subfigure}
\usepackage{latexsym}
\usepackage{tikz-qtree}
\usepackage{tabularx}
\usepackage{algorithm}
\usepackage{algorithmic}
\usepackage{paralist}
\usepackage{color,xcolor}
\usepackage[utf8]{inputenc}
\usepackage{nicefrac}       % compact symbols for 1/2, etc.
\usepackage{CJKutf8}
\usepackage{url}            % simple URL typesetting
\usepackage{booktabs}       % professional-quality tables
\usepackage{amsfonts}       % blackboard math symbols
\usepackage{lineno}
\usepackage{tikz}
\usepackage{mathtools}
\usepackage{helvet}
\usepackage{courier}
\usepackage{amsmath,mathrsfs,bm}
\usepackage{wrapfig}
\usepackage{bbding}
\usepackage{pifont}
\usepackage{caption}
\usepackage{arydshln}
\usepackage{colortbl}
\usepackage{float} 
\usepackage{amsthm,amssymb}
\usepackage{enumerate}
\usepackage{lscape}

\def\tsc#1{\csdef{#1}{\textsc{\lowercase{#1}}\xspace}}
\tsc{WGM}
\tsc{QE}
\tsc{EP}
\tsc{PMS}
\tsc{BEC}
\tsc{DE}
\newcommand{\paratitle}[1]{\vspace{1.4ex}\noindent \textbf{#1}}

\definecolor{myblue}{RGB}{157,224,255}
\definecolor{myred}{RGB}{255,180,219}
\definecolor{mygreen}{RGB}{167,252,211}

\begin{document}
\let\WriteBookmarks\relax
\def\floatpagepagefraction{1}
\def\textpagefraction{.001}

\begin{CJK}{UTF8}{gbsn}

\shorttitle{Grammar Induction from Visual, Speech and Text}

\shortauthors{Zhao et~al.}

\title[mode = title]{Grammar Induction from Visual, Speech and Text}

\author[hit]{Yu Zhao}
\ead{zhaoyucs@tju.edu.cn}
\credit{Programming, Writing \& Editing}

\author[nus]{Hao Fei}\corref{cor1}
\ead{haofei37@nus.edu.sg}
\credit{Conceptualization, Writing \& Editing}

\author[nus]{Shengqiong Wu}
\ead{swu@u.nus.edu}
\credit{Experiments, Writing \& Editing}

\author[hit]{Meishan Zhang}
\ead{zhangmeishan@hit.edu.cn}
\credit{Methodology, Review \& Editing}

\author[hit]{Min Zhang}
\ead{zhangmin2021@hit.edu.cn}
\credit{Discussion \& Review}

\author[nus]{Tat-seng Chua}
\ead{dcscts@nus.edu.sg}
\credit{Conceptualization, Review \& Editing}

\cortext[cor1]{Corresponding Author}

\affiliation[hit]{organization={Harbin Institute of Technology (Shenzhen)},
           city={Shenzhen},
           postcode={518055}, 
           % state={},
           country={China}}
\affiliation[nus]{organization={National University of Singapore},
           % addressline={}, 
           city={Singapore},
           postcode={118404}, 
           % state={},
           country={Singapore}}

\begin{abstract}
Grammar Induction (GI) seeks to uncover the underlying grammatical rules and linguistic patterns of a language, positioning it as a pivotal research topic within Artificial Intelligence (AI). Although extensive research in GI has predominantly focused on text or other singular modalities, we reveal that GI could significantly benefit from rich heterogeneous signals, such as text, vision, and acoustics.
In the process, features from distinct modalities essentially serve complementary roles to each other.
With such intuition, this work introduces a novel \emph{unsupervised visual-audio-text grammar induction} task (named \textbf{VAT-GI}), to induce the constituent grammar trees from parallel images, text, and speech inputs.
Inspired by the fact that language grammar natively exists beyond the texts, we argue that the text has not to be the predominant modality in grammar induction.
Thus we further introduce a \emph{textless} setting of VAT-GI, wherein the task solely relies on visual and auditory inputs.
To approach the task, we propose a visual-audio-text inside-outside recursive autoencoder (\textbf{VaTiora}) framework, which leverages rich modal-specific and complementary features for effective grammar parsing. 
Besides, a more challenging benchmark data is constructed to assess the generalization ability of VAT-GI system.
Experiments on two benchmark datasets demonstrate that our proposed VaTiora system is more effective in incorporating the various multimodal signals, and also presents new state-of-the-art performance of VAT-GI.
Further in-depth analyses are shown to gain a deep understanding of the VAT-GI task and how our VaTiora system advances.
Our code and data: {\small \url{https://github.com/LLLogen/VAT-GI/}}
\end{abstract}


\maketitle

\makeatletter
\def\old@comma{,}
\catcode`\,=13
\def,{%
  \ifmmode%
    \old@comma\discretionary{}{}{}%
  \else%
    \old@comma%
  \fi%
}
\makeatother

\section{Introduction}
% \vspace{-1mm}
Within the field of AI, human language acquisition is one of the important research topics. 
Human language knowledge involves various aspects, such as the vocabulary, phonetics, morphology, syntax, semantics and pragmatics of languages \cite{chomsky1957logical,osherson1995invitation,briscoe2000grammatical,flynn2012linguistic,fei2020retrofitting,fei2022lasuie,fei2021better,fei2024enhancing,DBLP:conf/acl/ChenLZZ24,jia2024llm}.
Among these, inferring the underlying grammatical rules and linguistic patterns of a language plays a significant role in language learning.
The process is also tasked as grammar induction \cite{DBLP:conf/naacl/KimRYKDM19,DBLP:conf/iclr/YogatamaBDGL17,DBLP:conf/iclr/ShenTSC19,DBLP:conf/eccv/HarwathRSCTG18}, aiming to uncover the latent structure of language constituents from natural inputs in an unsupervised manner.
In the community, the mainstream explorations on GI pay an extensive focus on the textual signals \cite{DBLP:conf/corr/abs-1904-02142,DBLP:conf/acl/KimDR19,DBLP:conf/conll/SpitkovskyAJM10,DBLP:conf/aaai/ChoiYL18,DBLP:conf/emnlp/HuMLM22,DBLP:conf/iclr/KimCEL20,DBLP:conf/emnlp/ZhaoT20,DBLP:conf/emnlp/ShiLG20}, as texts are the major medium of language.
However, it is important to note that human always acquire language knowledge with multimodal signals, i.e., text, vision and acoustic.
Knowing this fact, researchers carry out GI using different modalities of information, such as vision-language grammar learning \cite{DBLP:conf/emnlp/ZhaoT20,DBLP:conf/iccv/Hong0ZH21,DBLP:conf/iclr/WanHZT22,DBLP:conf/acl/ShiMGL19,DBLP:conf/coling/ShiMXJS18}, speech-based textless GI \cite{DBLP:journals/corr/abs-2310-07654} and video-aided GI \cite{DBLP:conf/naacl/ZhangSJXYL21,DBLP:conf/emnlp/ZhangSJM0YL22,DBLP:conf/icetc/Perdani22,DBLP:conf/aaai/PiergiovanniAR20}.

\begin{figure}[t]
% \vspace{-4.5mm}
\centering
\includegraphics[width=0.86\linewidth]{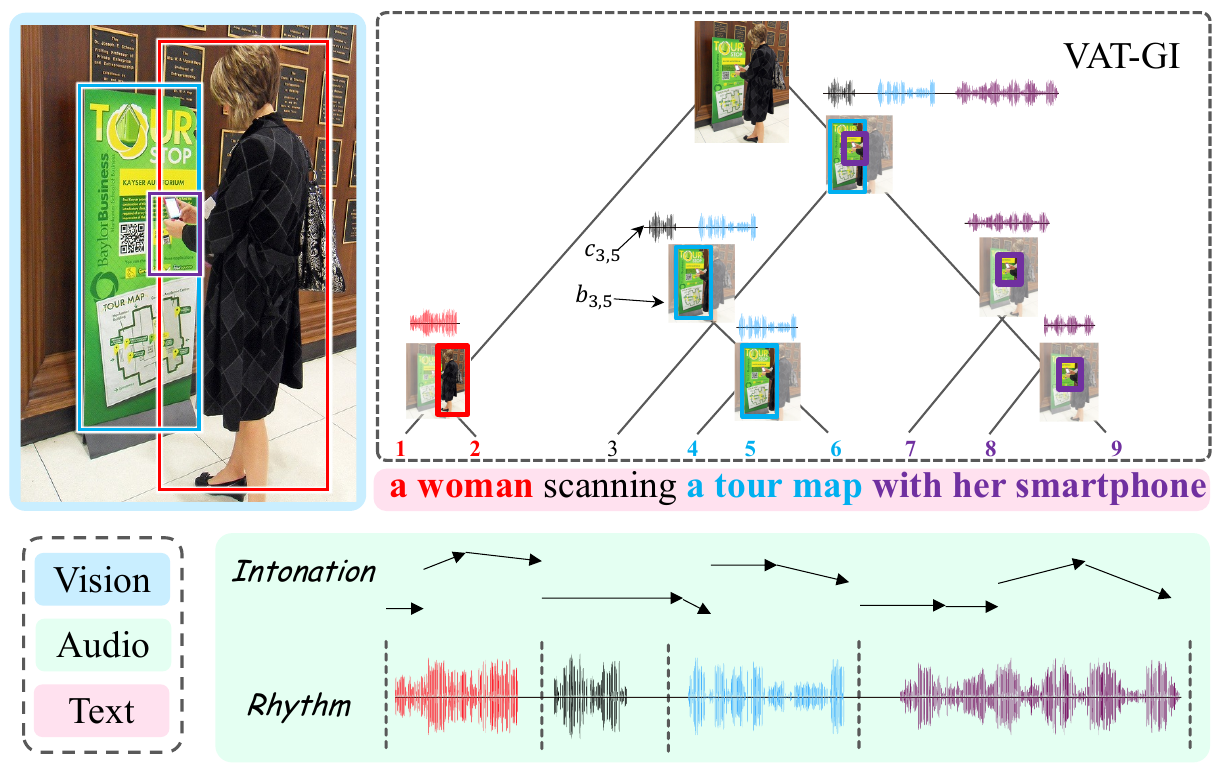}
% \vspace{-1.5mm}
\caption{
% Inducing the grammar
Unsupervised grammar induction with \textcolor{myblue}{vision}, \textcolor{mygreen}{audio} and \textcolor{myred}{text} modality sources, each of which contributes complementarily to the task.
% Best viewed by zooming in.
}
\label{fig:intro}
% \vspace{-4mm}
\end{figure}

Unfortunately, most existing researches on GI tend to overlook the mutually complementary contributions of the information in broad-covering modalities.
Actually, information and signals from all different modalities together can play crucial complementary roles in the acquisition of language throughout phylogenetic development. 
This can be exemplified by the cases in Figure \ref{fig:intro}:
the pixel regions in a visual image are often associated with the noun phrase structure, such as `\emph{a woman}' and `\emph{a tour map}',
while speech intonation and rhythm help segment sentences from a constituent-level perspective, such as `\emph{scanning a tour map}' and `\emph{with her smartphone}'.
This thus motivates us to introduce a novel task of unsupervised \emph{visual-audio-text grammar induction} (VAT-GI), with the aim of exploring the language learning of NLP-related AI systems in realistic scenarios.
Particularly, VAT-GI operates over parallel images, audios and texts, and extracts the shared latent constituency structures that encompass all these modalities (see $\S$\ref{Task Formulation} for task definition).
As depicted in Figure \ref{fig:intro}, the terminal nodes of the tree are the individual words, while non-terminal nodes are notional constituent spans, where nodes also associate with visual regions and speech clips.
Different sources of modalities together form an integral constituency grammar structure.

To effectively facilitate the multimodal grammar induction, a VAT-GI system should take into full consideration the distinct yet complementary characteristics of each modality:
% \vspace{-4.0mm}
\setdefaultleftmargin{2em}{2.2em}{1.87em}{1.7em}{1em}{1em}
\begin{compactitem}
    \item \textbf{Text:} Language intrinsically encompasses structural information by virtue of its inherent compositionality \cite{chomsky1957logical,mikolov2013distributed,DBLP:conf/iclr/ShenLHC18}, thereby providing a straightforward basis for grammar induction.
    
    \item \textbf{Vision:} The smaller spatial regions of pixels combine to form larger regions with higher-level visual semantics \cite{DBLP:conf/iccv/Hong0ZH21,DBLP:conf/emnlp/ZhaoT20,DBLP:conf/iclr/WanHZT22}, which correspond to the hierarchical structure observed among textual constituent spans.

    \item \textbf{Audio:} Speech conveys structural information through various intonation and rhythm patterns, naturally depicting the discontinuity and continuity of phonemes \cite{pickett1999acoustics,stockwell1960place,DBLP:journals/llc/ByunT16,DBLP:conf/sigmorphon/Muller02,DBLP:journals/llc/Heinz11a}.    
\end{compactitem}

% \textbf{Model Contribution.}
With the aforementioned observations, 
% To approach the VAT-GI task, 
in this work we present a nichetargeting framework to approach VAT-GI.
Built upon the deep inside-outside recursive autoencoder (Diora) model \cite{DBLP:conf/corr/abs-1904-02142} for effective text-based GI, we devise a novel \emph{visual-audio-text Diora} (namely \textbf{VaTiora}), which extends the Diora with the further capability of capturing specific structural characteristics of three modalities of sources, such as the compositionality of texts, the hierarchical spatial regions of images and the rhythm patterns of speech, and integrating them properly.
As shown in Figure \ref{fig:model}, first, rich modal-specific and complementary features, such as text embedding, visual regions and voice pitch, etc., are constructed from the input text, image and audio sources, respectively, all of which are effectively integrated into the inside-outside recursive autoencoding. 
Then, in the feature interaction phase, text embeddings are first fused with audio features via cross-attention to obtain the token-wise text-audio representations for constructing the span vectors; these span vectors are further mapped with fine-grained visual region features by an attention-aware fusion, thereby enhancing the span representations.
Moreover, the composition probabilities of visual regions are considered when calculating the span scores, during which the intonation pitch frequency features and voice activity features (representing the rhythm pattern) are also incorporated.
Through such cohesive integration, the cross-modal signals mutually reinforce each other for more accurate constituency parsing in the inside-outside pass of VaTiora.
We follow \cite{DBLP:conf/corr/abs-1904-02142} to train the VaTiora with structure reconstruction and contrastive learning.
In addition, we further introduce a cosine similarity learning objective across three types of features for representation alignment, bridging the heterogeneity among modalities.

Additionally, we introduce a \emph{textless} setting for VAT-GI, wherein the task solely relies on visual and auditory inputs, and the constituent tree is no longer structured around individual words but instead the segmented speech clips.
The motivation for setting the textless VAT-GI comes from the consideration of the languages that lack written forms, e.g., the languages of minority ethnic groups in certain regions.
We contend that text is not a fundamental modality for VAT-GI, as supported by the physical fact that \emph{language grammar natively exists beyond the texts}.
Through investigating the feasibility of textless VAT-GI, we uncover the potential of modeling syntactic structures using non-textual modalities.
Accordingly, we specially propose the aligned span-clip F1 metric (\textbf{SCF1}) to measure the textless VAT-GI, due to the inaccessibility of golden word segments of speeches in this setting.

VAT-GI can be evaluated on the public SpokenCOCO dataset \cite{DBLP:conf/acl/HsuHMSG20} with well-aligned images, texts and audio speeches.
However, SpokenCOCO has inherent limitations in terms of monotonous styles of visual scenes and speech tones, as well as short captions, which largely weakens the model generalization ability to complex signal inputs.
To combat that, we present a more challenging test set for VAT-GI, namely \textbf{SpokenStory}, which advances in richer visual scenes, timbre-rich audio and deeper constituent structures.
We collect 1,000 aligned image-text from Localized Narratives \cite{PontTuset_eccv2020}, and further record speech for each sentence by human speakers.
The experimental results on the datasets suggest that leveraging multiple modality features helps better grammar induction.
Further analyses demonstrate that VaTiora is effective in incorporating the various multimodal signals, and yielding new state-of-the-art induction performance of VAT-GI task.

In summary, this paper contributes in five key aspects.
\begin{compactitem}
    \item We for the first time introduce an important yet challenging task, unsupervised visual-audio-text grammar induction, to better emulate the human-level phylogenetic language acquisition;

    \item  We present a novel VAT-GI model, VaTiora, to properly navigate multimodal input sources;

    \item We newly contribute a challenging benchmark data, SpokenStory, for VAT-GI; and also propose a new metric for the textless VAT-GI evaluation;

    \item Our proposed method sets a strong-performing benchmark result for follow-up research. Our resources will be open to facilitate the community.
\end{compactitem}

The rest of the paper is organized as follows.
The next section contains the background and related work of textual grammar induction and multi-modal grammar induction. 
After that, in section 3 we present the definition visual-audio-text grammar induction task and how to evaluate it. 
In section 4, we elaborate on the details of the proposed VaTiora framework, which incorporates multiple cues from three modalities into an inside-outside algorithm.
Following that, in the experiment section, we compare our methods with strong grammar induction baselines, and explore how our method advances via in-depth quantitative and qualitative analysis.
Finally, we shed the light on the future work of this research, and then conclude the paper.

% \vspace{-2mm}
\section{Related Work}

In the field of AI, research draws inspiration from language to enhance the development of more intelligent systems and models. 
Among all topics, the task of grammar induction aims to infer the underlying grammatical rules and linguistic patterns of a language, thereby gaining a deeper understanding of the principles governing human intelligence.
GI unsupervisedly determines the constituent syntax of language, in the format of a phrase-structure tree, which intuitively depicts the language compositionality, i.e., how the small components (e.g., words or phrases) combine to larger constituents (e.g., clauses or sentences).
Such a process essentially emulates the human language acquisition.
Thus, GI \cite{lari1990estimation} has long been a fundamental topic in natural language processing (NLP) \cite{cohen2008shared,DBLP:conf/conll/SpitkovskyAJM10,DBLP:conf/acl/KleinM02,DBLP:conf/emnlp/HuMLM22,DBLP:conf/iclr/KimCEL20,DBLP:conf/emnlp/ZhaoT20,DBLP:conf/iccv/Hong0ZH21,DBLP:conf/iclr/WanHZT22,DBLP:conf/acl/ShiMGL19}.

According to the categories of information modalities, all existing GI research can be categorized into text-based GI, vision-language-based GI, and speech-based GI. 
Below, we survey relevant studies within each of these groups.

\paratitle{Textual Grammar Induction.}
The majority of previous GI studies pay the focus on the language domain with textual corpora \cite{cohen2008shared,lari1990estimation,DBLP:conf/conll/SpitkovskyAJM10,DBLP:conf/acl/KleinM02,DBLP:conf/emnlp/HuMLM22,DBLP:conf/iclr/KimCEL20}.
Generally, there are two types of frameworks for text-only grammar induction.
The first is PCFG-based methods, which assume the context-free grammar rules exist with a probability and estimates the likelihood of each latent rule \cite{DBLP:conf/acl/KimDR19,DBLP:conf/icgi/KuriharaS06,DBLP:conf/conll/WangB13}.
One way of inducing this probabilistic grammar is to fix the structure of a grammar and then find the
set of optimal parameters such that the resulting grammar best explains the language, which is usually approximated by a training corpus.
Early works induce PCFG via statistic-based methods, such as EM algorithm \cite{lari1990estimation}.
Following efforts focus on improving it by carefully-crafted auxiliary objectives \cite{DBLP:conf/acl/KleinM02}, priors or non-parametric models \cite{DBLP:conf/conll/WangB13}, and manually-engineered features \cite{DBLP:conf/acl/GollandDU12}.
\citet{DBLP:conf/acl/KimDR19} first propose to parameterize the PCFG's rule probability with a trainable neural network.
Jin et al. \cite{DBLP:conf/emnlp/JinDMSS18} applies depth bounds within a chart-based Bayesian PCFG inducer, limiting the search space of an unsupervised PCFG inducer.
The other type of framework is autoencoder-based methods, which estimate span composition likelihood without assuming context-free grammar.
\citet{DBLP:conf/corr/abs-1904-02142} first propose the Diora model, which incorporates the inside-outside algorithm into a latent tree chart parser.
% some research also explores the induction with other modalities of information.

\paratitle{Visual-Language Grammar Induction.}
The visually-grounded GI has received increasing research attention \cite{DBLP:conf/emnlp/ZhaoT20,DBLP:conf/iccv/Hong0ZH21,DBLP:conf/iclr/WanHZT22,DBLP:conf/acl/ShiMGL19}.
Since visual regions (e.g., objects of interest) intuitively encompass the correspondence of the textual spans (e.g., noun phrases), the visions are imported as a type of enhanced signal for better constituency tree parsing.
\citet{DBLP:conf/acl/ShiMGL19} first propose to import visual features into grammar induction, mapping visual and text via visual-semantic embedding.
The following works extend text-only GI models with extra visual information, such as VC-PCFG \cite{DBLP:conf/emnlp/ZhaoT20}, VG-NSL \cite{DBLP:conf/acl/ShiMGL19} and Cliora \cite{DBLP:conf/iclr/WanHZT22}.
The visually grounded compound PCFGs (VC-PCFG) extends the compound PCFG model (C-PCFG) by including a matching model between images and text.
Similar performance gain is also observed with VG-NSL.
Instead of fusing information of the entire image into the phrase representations, Cliora uses region-based fine-grained alignment to have a thorough understanding of the image.
There are also video-aided GI, where, compared with a single image, video frames can further facilitate the understanding of motion dynamics for the correspondence of textual predicates \cite{DBLP:conf/naacl/ZhangSJXYL21,DBLP:conf/emnlp/ZhangSJM0YL22,DBLP:conf/icetc/Perdani22,DBLP:conf/aaai/PiergiovanniAR20}.
Zhang et al. \cite{DBLP:conf/naacl/ZhangSJXYL21,DBLP:conf/emnlp/ZhangSJM0YL22} first propose video-aided GI models that prove videos can provide even richer information than images for grammar induction, including not only static objects but also actions and state changes useful for inducing verb phrases.

\paratitle{Speech-oriented Grammar Induction.}
On the other hand, acquiring languages from speech has also gained consistent research interests \cite{DBLP:conf/interspeech/KlasinasPIGM13,DBLP:journals/csl/IosifKAPGLP18,DBLP:conf/icassp/JansenDGJKCFHMRSCMVBBCDFHLLNPRST13,DBLP:conf/iclr/HarwathHG20,DBLP:conf/eccv/HarwathRSCTG18}, where the acoustic-prosodic features (e.g., phonetics, phonology) can offer important clues for the syntax induction from different perspectives than the visions and texts \cite{DBLP:journals/llc/ByunT16,DBLP:conf/sigmorphon/Muller02,DBLP:journals/llc/Heinz11a}.
Lai et al. \cite{DBLP:conf/asru/LaiSPKGCCBCHZLG23} first attempt incorporate audio signals for grammar induction and present the novel AV-NSL learner. 
They segment the speech waveform into sequences of word segments, and subsequently induce phrase structure using the inferred segment-level continuous representations.  
In this work, we take a combined holistic viewpoint, and investigate the GI under multimodal information sources, i.e., introducing a novel visual-audio-text grammar induction task.
This can be quite intuitive, as we humans always perceive the world with varied sensory inputs that can partially share the same common structures and meanwhile preserve distinctly complementary features, which together help achieve more effective GI.

Despite the significant achievements and extensive attention garnered by prior GI research, current approaches remain confined to single modalities, such as text or others, often overlooking the mutually complementary contributions of information from diverse modalities. 
In reality, information and signals from various sources—including text, vision, and acoustics—collectively and complementarily play crucial roles in language acquisition throughout phylogenetic development. 
Leveraging rich heterogeneous signals could greatly benefit GI. 
To address this, we propose a novel task, unsupervised visual-audio-text grammar induction (VAT-GI), to emulate real human-level phylogenetic language acquisition.

\section{Task Definition}
\paratitle{Formulation.}
\label{Task Formulation}
Given a sentence $X=\{x_1,x_2,...,x_n\}$ with $n$ words, an associated image $I$ and a piece of speech $S$ for $X$, the goal of unsupervised VAT-GI is to induce a constituency tree structure $T$ from $X$, $I$ and $S$ without any supervision of constituent structure annotations for training.
% , speech-text or region-text alignment annotations
As shown in Figure \ref{fig:intro}, the output tree structure $T$ is formed in a phrase constituent tree by Chomsky Normal Form (CNF) \cite{chomsky1959certain}, where each non-terminal node in the tree has exactly two children.
% $T$ is shared across all the three modalities, where 
Each node in $T$ contains a text span $(i,j), 1\leq i\leq j\leq n$, a region box $b_{i,j} \in \mathbb{R}^4$ in $I$ and a speech span $c_{i,j} \in \mathbb{R}^2$, where the text span $(i,j)$ is grounded to the region box $b_{i,j}$ and aligned to the speech span $c_{i,j}$.
% is aligned to $(i,j)$.
For the textless setting, the inputs of VAT-GI are only $I$ and $S$.
Thus the node set of output $T$ contains only $c_{i,j}$ and $b_{i,j}$.
\begin{figure}[t]
\centering
\includegraphics[width=0.8\linewidth]{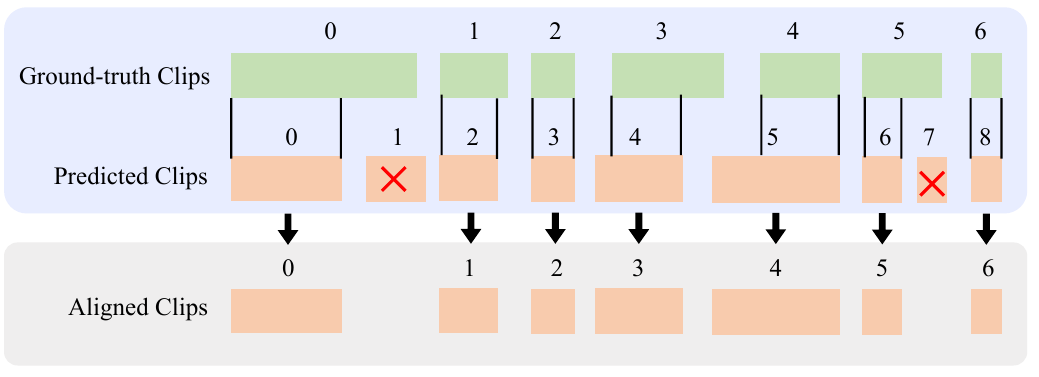}
\caption{
Illustration of clip alignment in SCF1.
}
\label{fig:metric}
\end{figure}

\paratitle{Evaluation.}
We adopt two widely-used metrics, \emph{averaged corpus-level F1 (Corpus-F1)} and \emph{averaged sentence-level F1 (Sent-F1)}, following \cite{DBLP:conf/acl/KimDR19}.
Corpus-level F1 calculates precision/recall at the corpus
level to obtain F1, while sentence-level F1 calculates F1 for
each sentence and averages across the corpus.
In this paper, we follow \cite{DBLP:conf/acl/KimDR19} discard trivial spans (length < 2) and evaluate on sentence-level F1 per recent work.
In textless settings, where there is no text script for speech-to-text alignment, we introduce a new metric, \emph{aligned span-clip F1 (SCF1)}, to measure the quality of the textless constituency tree.
Suppose the output of the textless VAT-GI contains a sequence of speech clips $\{clip^p_m\}$. 
The predicted constituency tree is denoted as the pairs of speech clips $\{c^p\}=\{(clip^p_h,clip^p_t)\}$, where each clip represents a token.
SCF1 first aligns predicted speech clips to ground-truth speech clips via tIoU (Temporal Intersection over Union) \cite{DBLP:conf/interspeech/PengH22}, which represents the overlap between two time intervals. It is a metric to measure the overlap ratio on the temporal dimension.
We use a greedy mapping from start to end, aligning predicted clip and ground-truth segmented clip if their tIoU is over a predefined threshold $p$.
In details, we traverse all the ground-truth clips, denoting $\{clip^g_n\}$, to search the aligned clips.
For each $clip^g_n$, we calculate the tIoU between $clip^g_n$ with all the predicted clips $\{clip^p_m\}$.
A predicted clip $clip^g_n$ is considered as the aligned one with $clip^p_m$ only when tIoU$(clip^g_n, clip^p_m)>p$.
% The tIoU \cite{DBLP:conf/interspeech/PengH22} is a metric to measure the overlap ratio on the temporal dimension.
If a ground-truth clip has multiple mapped predictions, we keep the one with the largest tIoU.
After alignment, we then calculate the F1 score based on TP, FP and FN.
We say the two spans are ``\textit{matched}'' when both their head and tail clips are aligned respectively.
Let $\{c^g\}=\{(clip^g_h,clip^g_t)\}$ be the ground-truth spans, then we have:
\begin{equation}%
\begin{split}
\text{TP} &= \text{Count}(pred \text{ and } gt \text{ are  matched}, pred \in \{c^p\} \text{ and } gt \in \{c^g\}), \\
\text{FP} &= \text{Count}(pred\in \{c^p\} \text{ that no } gt\in \{c^g\} \text{ could match}), \\
\text{FN} &= \text{Count}(gt\in \{c^g\} \text{ that no } pred\in \{c^p\} \text{ could match}),
\end{split}
\end{equation}
where $p$ is a threshold and Count($f$) denotes the number of samples that satisfies the condition $f$.
Figure \ref{fig:metric} gives the illustration of this mapping process.
% More details are given in Appendix \S\ref{app:c}.

\begin{figure*}[t]
\centering
\includegraphics[width=\linewidth]{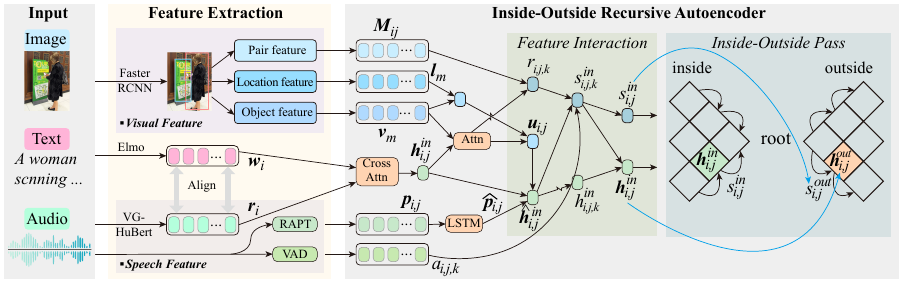}
% \vspace{-1mm}
\caption{
In our VaTiora framework,
\textbf{first} the feature extraction module constructs rich modal-specific features from the input image, text and speech.
RAPT: robust algorithm for pitch tracking; VAD: voice activity detection.
\textbf{Then} the inside-outside recursive autoencoder fuses various features and performs grammar induction.
}
\label{fig:model}
% \vspace{-3mm}
\end{figure*}

\section{Framework of VaTiora}
% \vspace{-1mm}
We propose a novel \textbf{v}isual-\textbf{a}udio-\textbf{t}ext \textbf{i}nside-\textbf{o}utside \textbf{r}ecursive \textbf{a}uto-\\encoder (dubbed VaTiora) framework based on the Diora model \cite{DBLP:conf/corr/abs-1904-02142}.
As shown in Figure \ref{fig:model}, VaTiora performs grammar induction with two key modules:
% consists of 
the multimodal feature extraction and the inside-outside recursive autoencoding.

% \vspace{-2mm}
\subsection{Feature Extraction}
\begin{figure}[t]
\centering
\captionof{table}{ 
Summary of all the features used in VaTiora.
}
\label{tab:feats}
\includegraphics[width=0.7\linewidth]{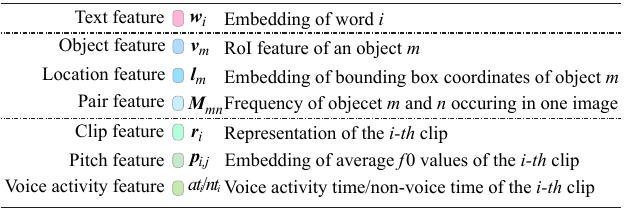}
\end{figure}

% \vspace{-1mm}
In the first step, we extract the modal-preserving features of each input modal from various perspectives, which are summarized in Table \ref{tab:feats}.

% \vspace{-2mm}
\paratitle{Textual Features.}
For text inputs, we follow the conventional GI methods, taking pre-trained word embeddings ELMo \cite{peterselmo-2018-naacl} to obtain the textual representations $\bm{W}=\{\bm{w}_1,...,\bm{w}_n\}$ from the input $X$.
In practice, we can also adopt other pre-trained word embeddings, such as Glove, or use randomly initialized embedding.

% \vspace{-2mm}
\paratitle{Visual Features.}
For visual feature extraction, given the input image $I$, we follow \cite{DBLP:conf/iclr/WanHZT22} to adopt an external object detector (Faster-RCNN \cite{girshick2015fast}) to extract a sequence of the object $\{o_1,...,o_M\}$ along with their object RoI features $\{\bm{v}_1,...,\bm{v}_M\}$.
Besides, each object $o_m$ has a bounding box $b_m$, which is encoded by an embedding layer to obtain the location features $\bm{l}_m$. Denote $b_m$ as:
\begin{equation}
b_m = (x_{min}, y_{min}, x_{max} ,y_{max}).    
\end{equation}
With the top-left as the origin point, we normalize it to:
\begin{equation}
    b'_m = (x_{min}/w, y_{min}/h, x_{max}/w, y_{max}/h),
\end{equation}
where $w,h$ is the width and length of the image.
The we embed $b'_m$ by:
\begin{equation}
    \bm{l}_m=\text{FC}(b_m). 
\end{equation}

We further consider the pair features of visual regions to determine the likelihood of object pairs forming a larger region.
Concretely, we maintain a pair relevance matrix $\bm{M} \in \mathbb{R}^{C \times C}$ for all categories of object pairs in the dataset based on the co-occurrence probability, where $C$ is the total number of object categories.
Supposing there are $C$ categories of object label, and for every two object that belong to $c_\alpha,c_\beta \in C$, their score is:
\begin{equation}\label{eq:1}
  \bm{M}_{\alpha,\beta} = \frac{\text{Co-Count}(\alpha, \beta)}{\sum_{\gamma\in C}(\text{Co-Count}(\alpha,\gamma)+\text{Co-Count}(\gamma,\beta))} \,,
\end{equation}
% \textcolor{red}{
where $\text{Co-Count}(\gamma,\beta)$ means the count that objects with categories $\alpha$ and $\beta$ detected in the same image within the entire dataset.

\paratitle{Speech Features.}
For speech features, we leverage a word discovery model VG-HuBert \cite{DBLP:conf/interspeech/PengH22} to segment raw speech into clips $\{c_1,...,c_T\}$, and obtain the clip representations $\{r_1,...,r_T\}$.
In the full setting, we take the textual transcripts to force align the speech, while in the textless setting, we use VG-HuBert to perform unsupervised word discovery.
Note that each clip corresponds to a single word in the sentence when text is provided as input, i.e., $T=n$.
% In the full setting, each word is aligned with a clip, resulting in aligned clips $\{c_i,...,c_n\}$ and $\{r_1,...,r_n\}$.
% as well as the word sequence.
To take into account the structural information of speech signals, we leverage pitch detection and voice activity detection (VAD) to extract intonation and rhythm features.
First, we adopt the robust algorithm for pitch tracking (RAPT) \cite{talkin1995robust}, extracting the $f0$ value for each speech clip $c_i$, and encode it to a high dimension representation.
Moreover, considering the pitch frequency constraint of normal human speech, we limit the extracted $f0$ value to the range of 50Hz to 500Hz \cite{trollinger2003relationships}.
Then, we embed the rounded $f0$ (450 values) to the high-dimension representation, and use an LSTM model to capture the temporal features of pitch changing:
\begin{equation}\label{eq:pitch}
\begin{split}
  f0_i &= \text{Round}(\text{Avg}(\text{RAPT}(c_i))), \\
  \bm{p}_i &= \text{Embed}(f0_i), \quad
  \hat{\bm{p}}_i = \text{LSTM}(\bm{p}_i),
\end{split}
\end{equation}
where 
% $c_i$ denotes the raw speech clip of the $i$-th word. 
Avg denotes average operation, Round means rounding to an integer. 
$f0_i$ means the average $f0$ of frames in clip $c_i$. 
$\hat{\bm{p}}_i$ represents the pitch feature and will be used to enhance span representation.
At last, we represent the rhythm feature of speech.
The voice activity time can be extracted by the robust voice activity detection (rVAD) method \cite{DBLP:journals/csl/TanSD20}.
For each clip, VAD outputs the number of speech frames and the non-speech frames.
We sum the frame length of speech frames and non-speech frames as voice activity time $at_i$ and non-voice time $wt_i$ for each clip $c_i$.

\subsection{Inside-Outside Recursive Autoencoder}
Similar to Diora, VaTiora operates as an autoencoder, encoding the sentence into span representations and then reconstructing the bottom words of the constituent tree, i.e., the terminal nodes.
The encoding process involves mapping the input words, denoted as $X$, to a latent constituent tree.
To efficiently explore all valid trees, a dynamic programming-based inside-outside algorithm \cite{lari1990estimation} is employed.
Notably, VaTiora incorporates external multi-modal cues during the inside-outside pass, thereby extending the functionality of the Diora framework.
Figure \ref{fig:model} provides an overview of the interaction of features and the inside-outside pass.

\paratitle{Feature Interaction.}
We first initialize the bottom-most terminal nodes\footnote{In the textless settings, we initialize $\bm{h}_{i,i}^{in}$ with only segmented speech clips $\bm{r}_i$, i.e., $\bm{h}_{i,i}^{in}=\text{Norm}(\bm{r}_i)$.} $\bm{h}_{i,i}^{in}$ with the aligned word embedding $\bm{W}=\{\bm{w}_i\}$ and speech clip representation $\bm{R}=\{\bm{r}_i\}$ via cross-attention:
\begin{equation}\label{eq:3}
\begin{split}
  &\{\bm{h}_{i,i}^{in}\} =\text{CrossAttn}(\bm{W},\bm{R})\,,\\
  &\text{CrossAttn}(\bm{W},\bm{R}) = \text{Softmax}((\bm{QW})(\bm{KR}))\bm{VR},
\end{split}
\end{equation}
where the $\bm{Q}$, $\bm{K}$, $\bm{V}$ are training parameters.
The bottom-most node span score $s_{i,i}^{in}$ is initialized with 0.
Similar to Diora, the VaTiora maintains a $N\times N$ chart $\bm{T}$ to store intermediate span vectors and scores, i.e., $\bm{h}_{i,j}^{in}$, $\bm{h}_{i,j}^{out}$ and $s_{i,j}^{in}$, $s_{i,j}^{out}$, for inside and outside representation respectively.
We further fuse the visual and speech features into span vector $\bm{h}_{i,j}^{in}$ and score $s_{i,j}^{in}$.
% First, for a decomposition $(i,k,j)$, 
First, for each span $(i,j)$ and its decomposition $(i,k,j)$,
we enhance $\bm{h}_{i,j}^{in}$ by:
\begin{equation}\label{eq:4}
\begin{split}
  &\hat{\bm{h}}_{i,k}^{in} = \bm{h}_{i,k}^{in} + \gamma \bm{u}_{i,k}  + \lambda \hat{\bm{p}}_{i,j}, \\
  &\bm{h}_{i,j,k}^{in} = \text{MLP}(\hat{\bm{h}}_{i,k}^{in};\hat{\bm{h}}_{k+1,j}^{in}),
  % \quad
  % \bm{h}_{i,j}^{in} = \sum_k
  \\
  &\bm{h}_{i,j}^{in} = \sum_k\bm{h}_{i,j,k}^{in}\cdot \mathop{\text{Softmax}}_k(s_{i,j,k}^{in}),
\end{split}
\end{equation}
% We use the $N\times N$ chart $T$ to store the intermediate span vectors and scores, i.e., $\bm{h}_{i,j}^{in}$, $\bm{h}_{i,j}^{out}$ and $s_{i,j}^{in}$, $s_{i,j}^{out}$.
% For inside pass, the model iteratively compute $\bm{h}_{i,j}^{in}$ and $s_{i,j}^{in}$ from bottom to up.
% We first compute a intermediate vector for each decomposition $(i,k,j)$
% :
where $1\leq i<k<j\leq N$, and $(;)$ denotes the concatenation.
% $f$ is a function of MLP network, 
$\gamma$ and $\lambda$ are the fusion weights.
The MLP (Multi-Layer Perceptron) is the network contains multiple linear layers, i.e., $\bm{h}_{l+1}=\sigma(\bm{W}_l\bm{h}_l+\bm{b}_l)$, where $\sigma$ is an activation function, $\bm{W}_l$ and $\bm{b}_l$ are training parameters.
We adopt a two-layer MLP here, which is used as the composition function to merge two spans (see Diora \cite{DBLP:conf/corr/abs-1904-02142}).
The $s_{i,j,k}^{in}$ is the decomposition score calculated during the inside pass (Eq. \ref{eq:6}).
Then $\bm{h}_{i,j}^{in}$ could be weighted sum of $\bm{h}_{i,j,k}^{in}$.
$\bm{u}_{i,j}$ and $\bm{p}_{i,j}$ are the visual and pitch feature, obtained by:
\begin{equation}\label{eq:5}
\begin{split}
  \bm{u}_{i,j} &= \sum_{m}^{M} (\bm{v}_m\textcolor{red}{+}\bm{l}_m)\cdot\text{Softmax}(\bm{v}_m^\top\bm{h}_{i,j}^{in}) ,
  % \text{Attn}_{i,j,m}\cdot(\bm{v}_m\oplus\bm{l}_m),\quad  \text{Attn}_{i,j,m}=\mathop{\text{softmax}}_m(\bm{v}_m^\top\bm{h}_{i,j}^{in}),\quad
  \\
  \hat{\bm{p}}_{i,j} &= \mathop{\text{AvgPool}}_{k\in[i,j]}(\hat{\bm{p}}_k),
\end{split}
\end{equation}
where $\bm{v}_m$ is the object-level region features and $\bm{l}_m$ is the embedding of the bouding box coordinates $b_m$.
$\hat{\bm{p}}_k$ is the pitch feature obtained in Eq. \eqref{eq:pitch}.
AvgPool($\cdot$) is the vector averaging operation.
% $\oplus$ denotes element-wise plus.

\begin{figure}[t]
\centering
\includegraphics[width=0.5\linewidth]{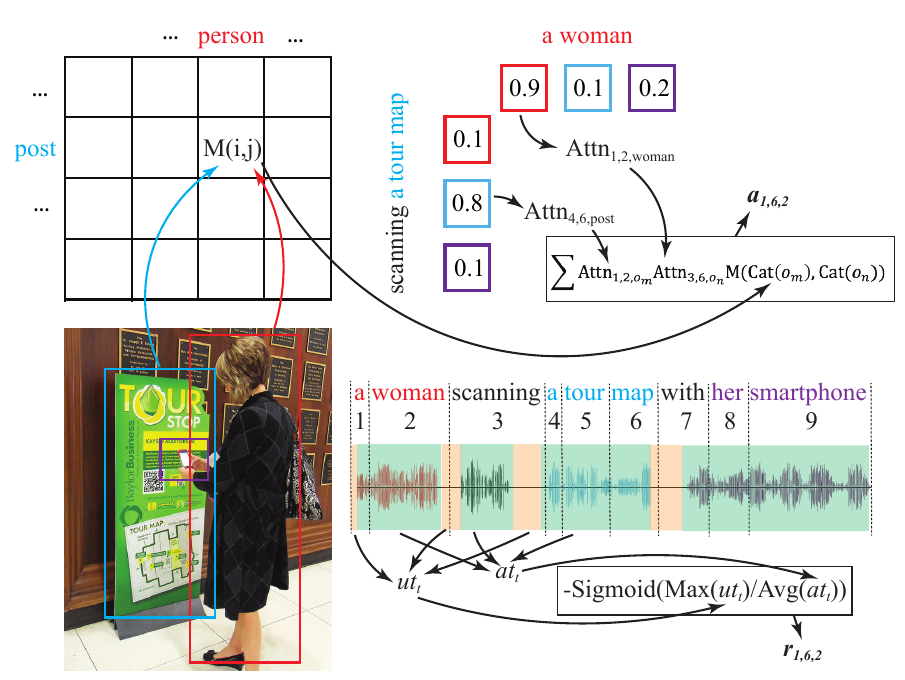}
\caption{
% Inducing the grammar
Illustration of pair feature and voice activity feature. 
}
\label{fig:app_ra1}
\end{figure}
For calculating the span score $s_{i,j}^{in}$, we consider fusing visual region composition and voice activity features.
Specifically, for each span $(i,j)$, we take its decomposition $(i,k,j)$ into account, where $1\leq i<k<j\leq N$.
Technically, we calculate a decomposition score $s_{i,j,k}^{in}$ with visual region composition score $r_{i,j,k}$ and the voice activity time-based decomposition score $a_{i,j,k}$.
For $r_{i,j,k}$, we compute:
\begin{equation}
\begin{split}
  r_{i,j,k} = \sum_{o_m,o_n}\text{Attn}_{i,k,o_m}&\cdot\text{Attn}_{k+1,j,o_n}\cdot \bm{M}_{\text{Cat}(o_m),\text{Cat}(o_n)}, \\
  \text{Attn}_{i,k,o_m} &= \text{Softmax}(\bm{v}_m^{\top}\bm{h}_{i,k}^{in}),\\
  \text{Attn}_{k+1,j,o_n} &= \text{Softmax}(\bm{v}_n^{\top}\bm{h}_{k+1,j}^{in})
\end{split}
\end{equation}
where $\bm{M}$ is the pair relevance matrix in Eq. \eqref{eq:1} and $\text{Cat}(*)$ retrieve the category index of an object.
% $o_m$.
For $a_{i,j,k}$, we compute:
\begin{equation}
  a_{i,j,k} = -\text{Sigmoid}(\text{Max}(ut_t) / \text{Avg}(at_t)), \; t\in[i,j)
\end{equation}
where $ut_t$ and $at_t$ are the non-voice time and voice activity time of speech clip $t$.
$\text{Max}$ is the maximum functions.
$a_{i,j,k}$ represents the density degree of the span $(i,j)$ in speech rhythm view.
Then the $s_{i,j,k}^{in}$ is computed as:
\begin{equation}\label{eq:6}
\begin{split}
  s_{i,j,k}^{in} = &(\hat{\bm{h}}_{i,k}^{in})^\top\hat{\bm{W}}_\theta(\hat{\bm{h}}_{k+1,j}^{in})+r_{i,j,k}\\
  &+a_{i,j,k} + s_{i,k}^{in} + s_{k+1,j}^{in},
\end{split}
\end{equation}
where the $\hat{\bm{W}}_\theta$ is the learnable parameter.
The final $s_{i,j}^{in}$ is obtained: 
\begin{equation}
    s_{i,j}^{in}=\sum_ks_{i,j,k}^{in}\mathop{\text{Softmax}}_k(s_{i,j,k}^{in}).
\end{equation}
Figure \ref{fig:app_ra1} visually illustrates the contributions of these features.

\begin{figure*}
\centering
\includegraphics[width=0.75\linewidth]{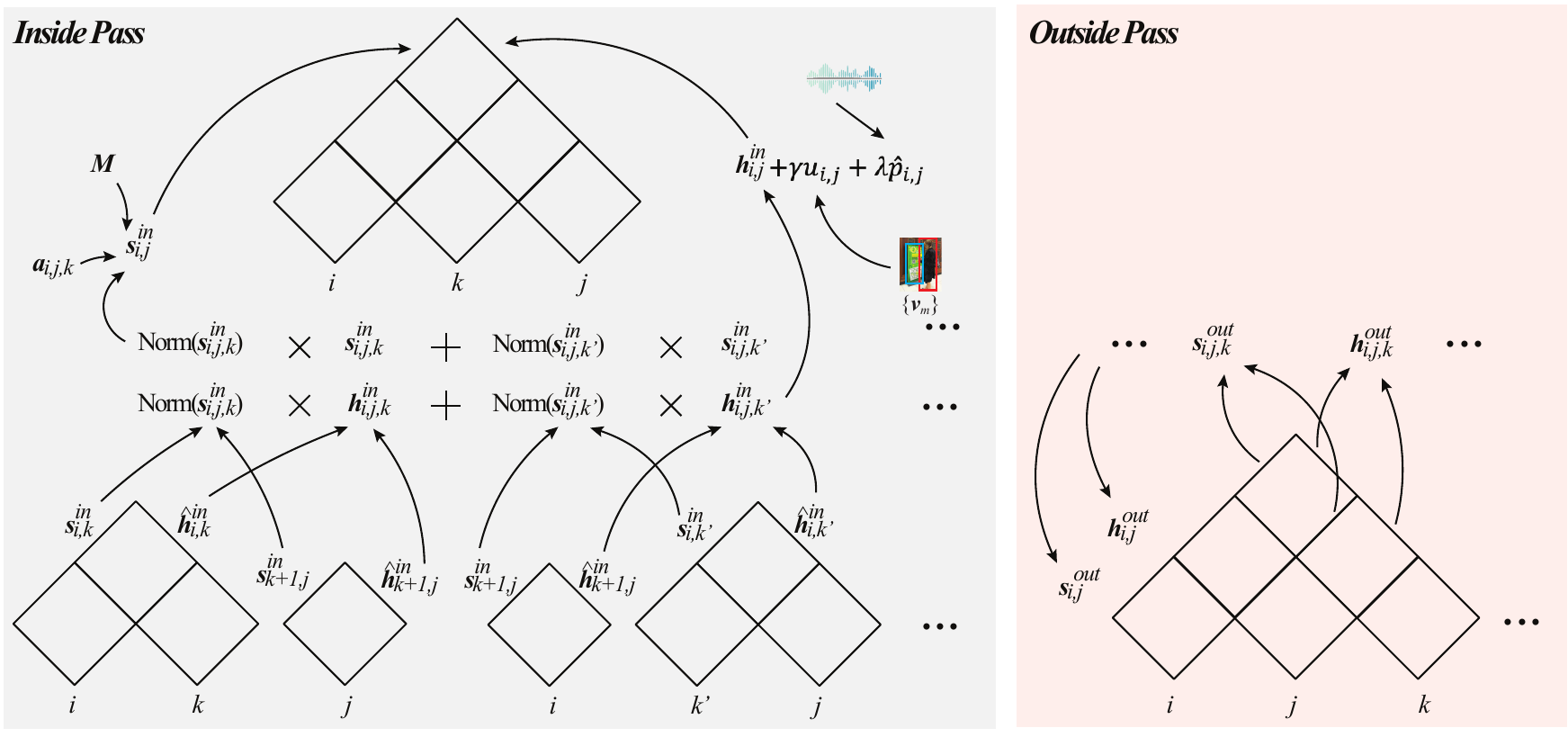}
\caption{
% Inducing the grammar
The process of inside and outside pass, respectively.
}
\label{fig:app_ra2}
\vspace{-4mm}
\end{figure*}
\paratitle{Inside-Outside Pass.}
The enhanced span vector $\bm{h}_{i,j}^{in}$ and score $s_{i,j}^{in}$ are then utilized to perform an inside-outside pass similar to Diora.
The process is based on the inside-outside algorithm \cite{baker1979trainable}, which is used for the derivation process of probabilistic context-free grammar.
This method produces representations for all internal nodes via recursively filling a chart, where each cell represents a node in the latent tree.

The inside pass of this method recursively compresses the input sequence, at each step inputting the vector representations of the two children into a composition function (i.e. Equation \ref{eq:4}) that outputs an inside vector representation of the parent.
This process continues up to the root of the tree, eventually yielding a single vector representing the entire sentence (Figure \ref{fig:app_ra2} left part).
In the inside pass, each span $(i,j)$ computes inside vector $\bm{h}_{i,j}^{in}$ and score $s_{i,j}^{in}$ by weighted summing all possible substructures.

Following that, the outside pass computes the outside presentations via a top-down process (Figure \ref{fig:app_ra2} right part).
The outside representations are encoded by looking at only the context of a given sub-tree.
The root node of the outside chart is learned as a bias.
Descendant cells are predicted using a disambiguation over the possible outside contexts.
Each component of the context consists of a sibling cell from the inside chart and a parent cell from the outside chart.
In the outside pass, we compute $\bm{h}_{i,j}^{out}$ and $s_{i,j}^{out}$ from top to bottom, where the bottom-most vector $\bm{h}_{i,i}^{out}$ is used to reconstruct the words.
For a span $(i,j)$ and a $k$ out of $(i,j)$:
\begin{equation}
    \bm{h}_{i,j,k}^{out}=\left\{
    \begin{aligned}
         \text{MLP}(\bm{h}_{i,k}^{out};\bm{h}_{j+1,k}^{in}) & \quad k>j \\
         \text{MLP}(\bm{h}_{k,j}^{out};\bm{h}_{k,i-1}^{in}) & \quad k<i
    \end{aligned}
    \right.
\end{equation}
\begin{equation}
    s_{i,j,k}^{out}=\left\{
    \begin{aligned}
         (\bm{h}_{i,k}^{out})^\top\bm{W}(\bm{h}_{j+1,k}^{in})\\+s_{i,k}^{out}+s_{j+1,k}^{in} & \quad k>j \\
         (\bm{h}_{k,j}^{out})^\top\bm{W}(\bm{h}_{k,i-1}^{in})\\+s_{k,j}^{out}+s_{k,i-1}^{in} & \quad k<i
    \end{aligned}
    \right.
\end{equation}
Similarly to the inside pass, the outside span representation $\bm{h}_{i,j}^{out}$ and score $s_{i,j}^{out}$ is:
\begin{equation}
\begin{split}
    \bm{h}_{i,j}^{out} &= \sum_k\bm{h}_{i,j,k}^{out}\cdot \mathop{\text{Softmax}}_k(s_{i,j,k}^{out}),\\
    s_{i,j}^{out} &= \sum_k s_{i,j,k}^{out}\cdot \mathop{\text{Softmax}}_k(s_{i,j,k}^{out}),
\end{split}
\end{equation}
Finally, the span score is calculated as $q(i,j)=s_{i,j}^{in}\cdot s_{i,j}^{out}/s_{1,n}^{in}$ to measure how likely the span $(i,j)$ exists.

% \vspace{-3mm}
\subsection{Overall Training}
% We process reconstruction, contractive learning and representation learning while training VaTiora.
% \vspace{-1mm}
\paratitle{Structure Reconstruction and Contrastive Learning.}
Following \cite{DBLP:conf/iclr/WanHZT22,DBLP:conf/corr/abs-1904-02142}, 
we adopt structure reconstruction and contrastive learning
% we use similar training objectives for structure reconstruction and contrastive learning.
For reconstruction, a self-supervised blank-filling objective is defined as:
\begin{equation}
  \mathcal{L}_{rec} = -\frac{1}{n}\sum_i\log P(x_i|\bm{h}_{i,i}^{out}).
\end{equation}
For contrastive learning, we randomly select unpaired image $I'$ and span $(i,j)'$ as negative samples within a training batch, and calculate the contrastive objective:
\begin{equation}
\begin{split}
  &l_{span}(I,i,j)=\max\{0,d(I,(i,j)')-d(I,(i,j))+\epsilon\} \\
  &\quad\quad\quad\quad +\max\{0,d(I',(i,j))-d(I,(i,j))+\epsilon\} \,,\\
  &d(I,(i,j))=sim((i,j),I)\times q(i,j), i\ne j,\\
  &sim(i,j,I)=\max_{m\in[0,M]}\{\bm{v}_m^\top(\bm{h}_{i,j}^{in}+\bm{h}_{i,j}^{out})\},
  % &sim(i,i,I)=\max_{m\in[0,M]}\{\bm{v}_m^\top\bm{h}_{i,i}^{in}\},
\end{split}
\end{equation}
\begin{figure}[t]
% \vspace{-2mm}
\centering
\includegraphics[width=0.5\linewidth]{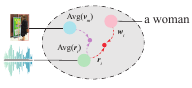}
% \vspace{-1.5mm}
\caption{
Illustration of the representation learning to map word-level speech clip $r_i$ and word embedding $w_i$, and whole image $\text{Avg}(\{\bm{v}_m\})$ and speech $\text{Avg}(\{\bm{r}_i\})$.
}
\label{fig:loss}
% \vspace{-4mm}
\end{figure}
where $\epsilon$ is the positive margin.
$q(i,j)$ is the span score as mentioned above ($q(i,j)=s_{i,j}^{in}\cdot_{i,j}^{out}/s_{1,n}^{in}$).
% where $d(I,(i,j))= \text{Max}(\bm{})\times q(i,j), i\ne j$ and $\epsilon$ is the positive margin.
% $I'$ and $(i,j)'$ means negative exmples.
For bottom-most $(i,i)$, e.g., the word $w_i$ (or the speech clip $c_i$ in textless setting), we compute:
\begin{equation}
  l_{word}(I,i) = -\log\frac{\exp(sim(i,I))}{\sum_{\hat{I}\in batch}\exp(sim(i,\hat{I}))},
\end{equation} 
where $sim(i,I)=\max_{m\in[0,M]}\{\bm{v}_{m}^\top\bm{h}_{i,i}\}$ 
% ($sim(i,I)=\max\{\bm{v}_m^\top\bm{r}_i\}$ in textless setting).
The final contrastive loss will be:
\begin{equation}
  \mathcal{L}_{cl}=\sum_{i,j, i\ne j} l_{span}(I,i,j)+\sum_{i} l_{word}(I,i).
\end{equation}

% \vspace{-3mm}
\paratitle{Representation Learning.}
Furthermore, we propose to use a representation learning objective to align the vectors of multi-modal inputs in the feature space:
\begin{equation}
\begin{split}
  &//\text{\em Full setting}\\
&\mathcal{L}_{rep} = \text{Cos}(\bm{r}_i, \bm{w}_i) + \text{Cos}(\text{Avg}(\{\bm{r}_i\}), \text{Avg}(\{\bm{v}_m\})),\\
&//\text{\em Textless setting}\\
&\mathcal{L}_{rep} = \text{Cos}(\text{Avg}(\{\bm{r}_i\}), \text{Avg}(\{\bm{v}_m\})),
\end{split}
\end{equation}
where Cos means cosine similarity. 
$\text{Avg}(\{\bm{r}_i\})$, $\text{Avg}(\{\bm{v}_m\})$ means average of the clip and object vectors, representing the whole speech and image feature. In textless setting, there are only $\text{Cos}(\text{Avg}(\{\bm{r}_i\}), \text{Avg}(\{\bm{v}_m\}))$.
The whole training loss will be:
% $\mathcal{L}= \mathcal{L}_{rec} + \alpha_1 \mathcal{L}_{cl} + \alpha_2 \mathcal{L}_{rep}$.
\begin{equation}
  \mathcal{L}= \mathcal{L}_{rec} + \alpha_1 \mathcal{L}_{cl} + \alpha_2 \mathcal{L}_{rep}.
\end{equation}

In the inference stage, we follow the conventional GI method to predict the tree with the maximum inside scores via the CKY algorithm.

\begin{figure*}[!t]
\centering
\includegraphics[width=\linewidth]{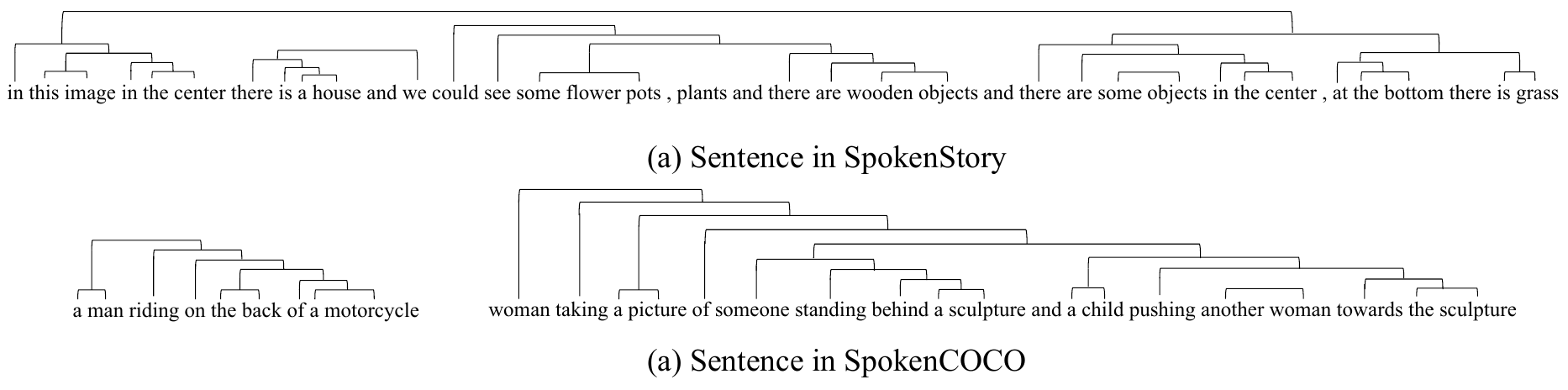}
\caption{
Comparison of captions between our SpokenStory and SpokenCOCO. 
}
\label{fig:app_sample}
\end{figure*}

\begin{figure*}[!t]
\centering
\includegraphics[width=0.9\linewidth]{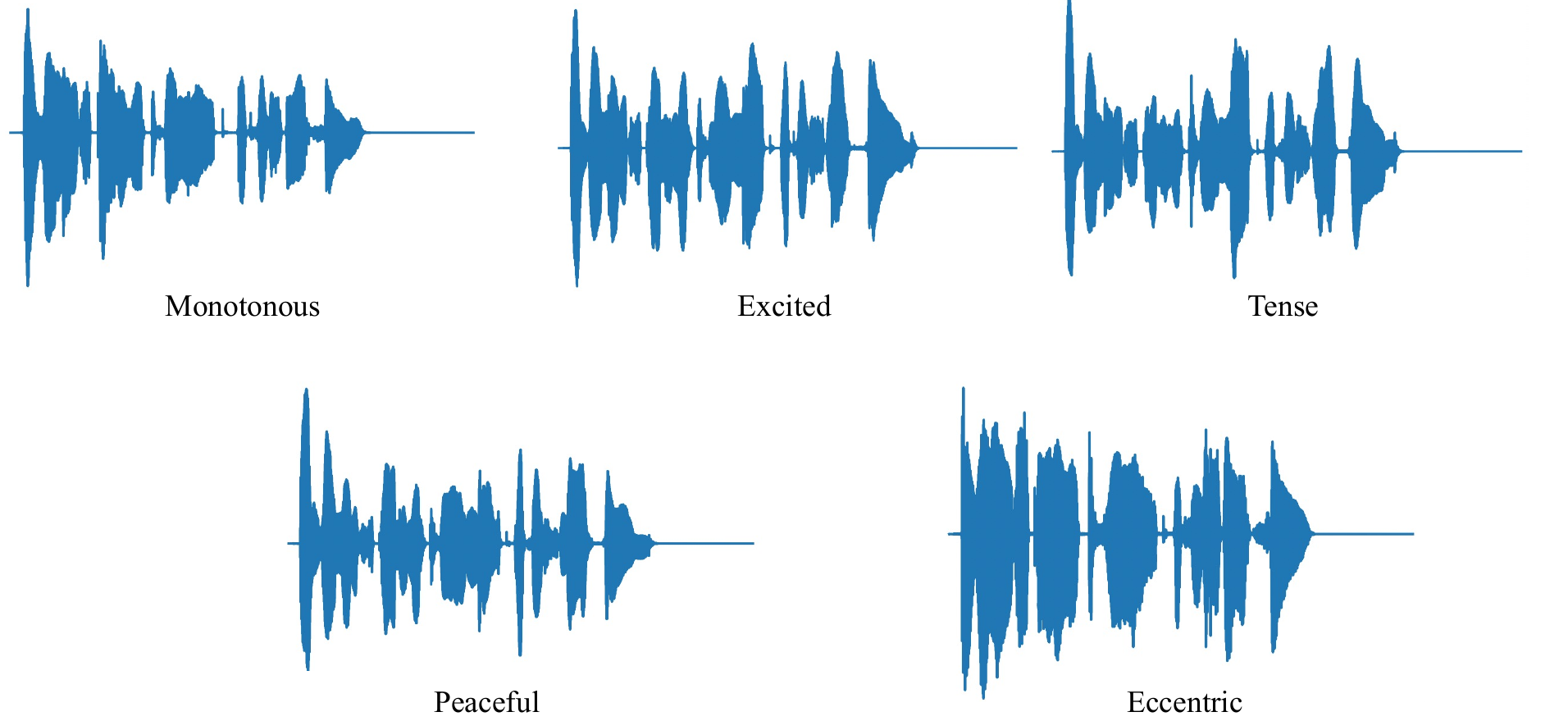}
\caption{
% Inducing the grammar
Visualized waveform for the sentence ``\emph{A woman scanning a tour map with her smartphone}'' in the five emotions. 
}
\label{fig:app_wav}
\end{figure*}
% the data statistics.
\begin{table}[t]
    \fontsize{10}{12}\selectfont
    \setlength{\tabcolsep}{2.5mm}
    
    \begin{center}
    % \resizebox{1\textwidth}{!}{
    % \vspace{-6.5mm}
    \caption{
    Statistics of SpokenCOCO and SpokenStory. 
    We follow \cite{DBLP:journals/pami/KarpathyF17} to split SpokenCOCO. ``Avg. Sppech Len.'' means average speech length. ``Agb. Sent. Len.'' means average sentence length.
    % ``$\star$'' means the number is for the whole OpenImage images.
    % SpokenStory dataset and SpokenCOCO test set.
    }
    \vspace{-2mm}
    \begin{tabular}{lccc}
    \hline
    \multicolumn{1}{c}{} & \small{\bf SpokenCOCO} & \small{\bf SpokenStory} \\
    \hline
    Split & Train/Val/Test & Test \\
    % \hline
    % \multicolumn{4}{l}{\textbf{2D vs. 3D Info}} \\
    % & 2D & - & - \\
    % & 3D & - & - \\
    Images & 80K/5K/5K & 1K \\
    Visual scenario & 91 & > 300 \\
    \cdashline{1-3}
    Avg. Speech Len. (second) & 2.54 & 5.01 \\
    Tone Styles & Monotonous & 5 Styles \\
    \cdashline{1-3}
    Avg. Sent. Len. (word) & 10.46 & 20.69 \\
    \hline
    \end{tabular}
    % }
    \label{tab:data}
    \end{center}
    \vspace{-5mm}
\end{table}

\section{Novel Dataset for VAT-GI}
VAT-GI can be evaluated on the SpokenCOCO dataset \cite{DBLP:conf/acl/HsuHMSG20}, which contains approximately 600,000 speech recordings of MSCOCO image captions\footnote{\url{https://cocodataset.org/}}.
Nevertheless, SpokenCOCO can fall prey to limitations in terms of its \emph{visual scenario styles, short captions}, and \emph{monotonous speech tone}, resulting in inferior generalization performance of VAT-GI parser across diverse language environments.
Thus, we introduce a more challenging test set, namely SpokenStory.
The dataset is built by extending the image caption data with speech records.
We comprises 1,000 images extracted from OpenImages \cite{OpenImages} across various scenarios, and accompanied by corresponding caption annotations from \cite{PontTuset_eccv2020}.
OpenImages is widely used for computer vision research and applications, which contains over 9 million labeled images sourced from the web and covers more than 6,000 topics and categories.
Each image is annotated with labels that describe the objects or concepts present in the image.

We adopt the annotations of Localized Narratives \cite{PontTuset_eccv2020} for aligned captions, and we manually record speech for each caption by human speakers, who are recruited in our research team, producing the aligned transcripts and speech audios.
The average sentence length of SpokenStory is 20 words, with an average speech duration of 5.01 seconds, roughly twice that of SpokenCOCO.
Figure \ref{fig:app_sample} shows the comparison of captions between our SpokenStory and SpokenCOCO, where the sentence in SpkenStory has longer constituents.
The speech recordings encompass five distinct tone styles, i.e., ``{\it Monotonous}'', ``{\it Excited}'', ``{\it Tense}'', ``{\it Peaceful}'' and ``{\it Eccentric}'', with each characterized by specific intonation and rhythm.
We do not suggest speakers how to express each emotion while letting them express all by their own comprehension.
Figure \ref{fig:app_wav} illustrates the waveform of five types of speech in the same sentence.
We develop a tool for recording and transcript annotation.
The speakers first record the captions and the system will return a pseudo transcript via an off-the-shelf speech segmentation model.
After recording, the speakers are asked to adjust the transcript to check the word boundaries.
Then the recording and the transcripts will be saved into the database.
Overall, the new test set poses a more challenging setting, due to its longer sentence, complex constituents and diverse intonations and tones.
% Details are shown in Appendix \S\ref{app:c}.
Table \ref{tab:data} compares two datasets.

% \vspace{-2mm}

\section{Experiments}
\subsection{Settings}
Following \cite{DBLP:conf/cvpr/ZhongYZLCLZDYLG22}, we use the Faster-RCNN model to detect object regions and extract visual features of the image.
For fair comparison, we follow \cite{DBLP:conf/iclr/WanHZT22} that use ELMo \cite{girshick2015fast} for text embedding.
Note the PLM or LLM embedding may contain potential knowledge of language structures during the pre-training stage, which may introduce immeasurable interference to the experiments.
Following previous work, the ground-truth constituent structures are parsed with the Benepar \cite{DBLP:conf/acl/KleinK18}, and the we take the hand-checked text set from Shi et al. \cite{DBLP:conf/acl/ShiMGL19}.
% We train our modal on SpokenCOCO train set and evaluate on both SpokenCOCO test set and SpokenS
We use VG-HuBert \cite{DBLP:conf/interspeech/PengH22} model for the speech encoding and word segmentation, which is pre-trained on unlabeled speech-image pairs.
For the inside-outside recursive autoencoder, we follow \cite{DBLP:conf/corr/abs-1904-02142} to use an MLP as the composition function for both inside and outside passes.
We compare our VaToria with current state-of-the-art methods on two kinds of settings: 1) text-only grammar induction. 2) Visual-text grammar induction on the same dataset, so that we could explicitly explore the influence of different modalities.
The model hyper-parameters are listed in Table \ref{tab:hyperparameter}. 
Other settings have defaulted to Diora.
% More implementation details and hyper-parameters are shown in Appendix \S\ref{app:c}.
\begin{table}[t]
    \fontsize{9}{11}\selectfont
    \setlength{\tabcolsep}{3.8mm}
    \begin{center}
    % \resizebox{1\textwidth}{!}{
    \caption{Model hyperparameters.}
    % \vspace{-3mm}
    \begin{tabular}{lclclc}
    \toprule
     \bf Hyper-param. & \bf Value & \bf Hyper-param. & \bf Value \\
     
    \midrule
    dimension of ROI feature & 2048 & $\alpha_2$ & 0.5 \\
    dimension of Speech hiddens & 1024 & max text length & 80 \\
    dimention of word embeddings & 400 & optimizer & Adam\\
    number of RoI & 36 & dropout & 0.1\\
    tIoU threshould of SCF1 & 0.5 & learning rate & 1e-4\\
    $\gamma$ & 0.5 & bach size & 64 \\
    $\lambda$ & 0.5 & epoch & 30 \\
    $\alpha_1$ & 0.5 \\
    
    % $\alpha_2$ & 0.5 \\
    % max text length & 80 \\
    
    % optimizer & Adam \\
    % dropout & 0.1 \\
    % learning rate & 1e-4 \\
    % bach size & 64 \\
    % epoch & 30 \\
    \bottomrule
    \end{tabular}
    % }
    \label{tab:hyperparameter}
    \end{center}
    \end{table}

\subsection{Baseline Specification}
We compare our model with baselines in two types of settings:
\paragraph{Text Only Methods,} which only takes text as inputs.
\begin{itemize}
    \item {\bf Left Branching}, A rule-based construction method that compose each span with its next word to a higher span from left to right.
    \item {\bf Right Branching}, A rule-based construction method that compose each span with its prior word to a higher span from right to left.
    \item {\bf Random}, Randomly composing spans to a tree.
    \item {\bf C-PCFG}, A method based on a compound probabilistic context-free grammar, where the grammar’s rule probabilities are modulated by a per-sentence continuous latent variable.
    \item {\bf Diora}, A deep inside-outside recursive autoencoder, which incorporates the inside-outside algorithm to compute the constituents and construct the constituency structure.
    This is also the backbone of our framework.
\end{itemize}

\paragraph{Visual-Text Methods,} which takes aligned image and description text as inputs.
\begin{itemize}
    \item {\bf VG-NSL}, A visually grounded neural syntax learner that learns a parser from aligned image-sentence pairs. The model is optimized via REINFORCE, where the reward is computed by scoring the alignment of images and constituents.
    \item {\bf VG-NSL+HI}, VG-NSL model with the design of head-directionality inductive biases, encouraging the model to associate abstract words with the succeeding constituents instead of the preceding ones.
    \item {\bf VC-PCFG}, An extended model of C-PCFG with visually grounded learning.
    \item {\bf Cliora}, An extended model of Diora with a fine-grained visual-text alignment via mapping RoI regions and spans. The model computes a matching score between each constituent and image region, trained via contrastive learning.
\end{itemize}

\subsection{Significance Test}
We employ five different random seeds to initialize the model parameters and the data shuffling to obtain a set of results, which are used for significance test.
We apply two-tailed T-test \cite{pillemer1991one} for the significance and use the $p$ value to measure the significance levels of all our results.
The significance levels are reported in each table via a postfix tag, where `$\uparrow$', `$\uparrow\uparrow$', `$\uparrow\uparrow\uparrow$' denote $p<0.1$, $p<0.05$, and $p<0.01$. 
The `-' denote that the results are not significant or lower than the baselines.

\begin{table}[t]
    % \fontsize{8}{10}\selectfont
    \fontsize{10}{12}\selectfont
    \setlength{\tabcolsep}{6mm}    
    \begin{center}
    % \resizebox{1\textwidth}{!}{
    % \vspace{-1.5mm}
    \caption{
    Main results of Corpus-F1 and Sent-F1 on the SpokenCOCO test set. In the VAT-GI-Textless setting, we report Corpus-SCF1 and Sent-SCF1. The up arrows represent the significance level of our model to the baselines.
    }\label{tab:main_results}
    \begin{tabular}{lccc}
    \hline
    \multicolumn{1}{c}{} & \bf Corpus-F1 & \bf Sent-F1\\
    % \hline
    % \multicolumn{4}{l}{\textbf{2D vs. 3D Info}} \\
    % & 2D & - & - \\
    % & 3D & - & - \\
    % Only Language
    \hline
    \multicolumn{3}{l}{$\bullet$ \textbf{Text-only GI}} \\
    \quad Left Branch & 15.1($\uparrow\uparrow\uparrow$) & 15.7($\uparrow\uparrow\uparrow$)\\
    \quad Right Branch & 51.0$\uparrow\uparrow\uparrow$ & 51.8($\uparrow\uparrow\uparrow$)\\
    \quad Random & 24.2($\uparrow\uparrow\uparrow$) & 24.6($\uparrow\uparrow\uparrow$)\\
    \quad C-PCFG & 53.6($\uparrow\uparrow\uparrow$) & 53.7($\uparrow\uparrow\uparrow$)\\
    \quad Diora & 58.3($\uparrow\uparrow\uparrow$) & 59.1($\uparrow\uparrow\uparrow$)\\
    \cdashline{1-3}
    % \hdashline
    % % Vision-language 
    \multicolumn{3}{l}{$\bullet$ \textbf{Visual-Text GI}} \\
    \quad VG-NSL & 50.4($\uparrow\uparrow\uparrow$) & -\\
    \quad VG-NSL+HI & 53.3($\uparrow\uparrow\uparrow$) & -\\
    \quad VC-PCFG & 59.3($\uparrow\uparrow\uparrow$) & 59.4($\uparrow\uparrow\uparrow$)\\
    \quad Cliora & 61.0($\uparrow\uparrow$) & 61.7($\uparrow\uparrow\uparrow$)\\
    % \hdashline
    \cdashline{1-3}
    \multicolumn{3}{l}{$\bullet$ \textbf{VAT-GI}} \\
    \quad VaTiora & \bf 62.7 & \bf 63.0\\
    \hline\hline
    \multicolumn{1}{c}{} & \bf Corpus-SCF1 & \bf Sent-SCF1\\
    % \hdashline
    % \cdashline{1-3}
    \hline
    \multicolumn{3}{l}{$\bullet$ \textbf{VAT-GI-Textless}} \\
    % \quad VaTiora (Textless) & 32.45 & 31.19\\    
    \quad VaTiora & 32.5 & 31.2\\

    \hline
    \end{tabular}
    % }
    \end{center}
    % \vspace{-3mm}
\end{table}
\begin{table}[!pt]
    % \fontsize{8}{10}\selectfont
    \fontsize{10}{12}\selectfont
    \setlength{\tabcolsep}{3.mm}
    \begin{center}
    % \resizebox{1\textwidth}{!}{
    % \vspace{-6mm}
    % \vspace{-2mm}
    \caption{
    Corpus-F1 on the SpokenCOCO test set and SpokenStory set. ``AvgSentLen'' means the average sentence length. All the models are trained on the SpokenCOCO train set.
    }
    \label{tab:res_story}
    \begin{tabular}{lccc}
    \hline
    \multicolumn{1}{c}{} & \bf SpokenCOCO & \bf SpokenStory\\
    \hline
    $\bullet$ \textbf{AvgSentLen} & 10.46 & 20.69 \\
    \hline
    \quad Diora & 58.3($\uparrow\uparrow\uparrow$) & 35.74($\uparrow\uparrow\uparrow$) \\
    \quad Cliora & 61.0($\uparrow\uparrow$) & 37.82($\uparrow\uparrow\uparrow$) \\
    \quad VaTiora & \bf 62.70 & \bf 53.19 \\
    \hline
    \end{tabular}
    % }
    \end{center}
    \vspace{-3mm}
\end{table}
\subsection{Main Results}
% \vspace{-1mm}
As shown in Table \ref{tab:main_results}, we compare our VAT-GI with two settings, i.e., text-only and visual-text.
When comparing the PCFG-based methods (i.e., C-PCFG and VC-PCFG) and Diora-based methods (i.e., Diora and Cliora) separately, we can see the models in visual-text setting perform better, demonstrating that the visual information benefits the GI.
Overall, the Diora-based methods outperform the other methods.
% which is consistent with the reports in other datasets \cite{DBLP:conf/iclr/WanHZT22}.
When extending to VAT-GI, our model outperforms the state-of-the-art Cliora by leveraging both visual and speech information, demonstrating the multiple modality features helps better grammar induction.
In the textless setting, the performance declines sharply, which is reasonable due to the lack of text-level structure.
But we can still learn the latent structure by speech and outperforms the random setting.
We also evaluate our model on the new proposed dataset SpokenStory.
% The SpokenStory is a test set, and we evaluate all the models trained on the SpokenCOCO train set.
For a fair comparison, we report the three Diora-based methods trained on the SpokenCOCO train set.
The results are presented in Table \ref{tab:res_story}.
Among these methods, our VaTiora demonstrates significant improvements, particularly in the more challenging SpokenStory dataset,
% As shown in Table \ref{tab:res_story}, our VaTiora shows larger improvements in the harder SpokenStory,
gaining about 15.37 Corpus-F1 over the current best-performing model, Cliora.
The results largely indicate the potential of VaTiora in effectively handling complex constituent structures.
% has great promise in handling complicated constituent structures.

\begin{table}[t]
    % \fontsize{8}{10}\selectfont
    \fontsize{10}{12}\selectfont
    \setlength{\tabcolsep}{3mm}
    
    \begin{center}
    % \resizebox{1\textwidth}{!}{
    % \vspace{-8mm}
    % \vspace{-2mm}
    \caption{
    Module ablation results. ``Pitch'' and ``VA' denote pitch feature and voice activity feature, i.e., $\bm{p}_i$ and $a_{i,j,k}$. ``w/o Vision'' means ablating all the visual features. ``Region'' denotes region features, ablating which means using the feature of the whole image instead.
    ``Pair'' means region pair score, i.e., $\bm{M}$.
    % ``w/o crossAttn'' means replace cross-attention in Eq. \ref{eq:3} with element-wise plus. ``w/o LSTM'' means replace $\bm{p}_{i,j}$ with $\hat{\bm{p}}_{i,j}$ in Eq. \ref{eq:5}.
    }
    \label{tab:ablation}
    \begin{tabular}{cccccccc}
    \hline
    \bf Speech & \bf Pitch & \bf VA & \bf Visual & \bf Region & \bf Pair & \bf Corpus-F1\\
    \hline
    \checkmark & \checkmark & \checkmark & \checkmark & \checkmark & \checkmark & 62.70 \\
    
    \checkmark & \checkmark & \ding{55} & \checkmark & \checkmark & \checkmark & 62.23 \\
    \checkmark & \ding{55} & \checkmark & \checkmark & \checkmark & \checkmark & 62.10 \\
    \checkmark & \ding{55} & \ding{55} & \checkmark & \checkmark & \checkmark & 61.38 \\
    \ding{55} & - & - & \checkmark & \checkmark & \checkmark & 60.07 \\

    % \hline
    % \multicolumn{1}{l}{\quad w/o Speech} & & & 60.07 \\
    % \cmidrule(r){2-3}
    % \multirow{4}{*}{\quad w/ ~~Speech} & Pitch & VA &\\
    % % \cmidrule(r){2-3}
    % & \ding{55} & \ding{55} & 61.38\\
    % & \checkmark & \ding{55} & 62.23\\
    % & \ding{55} & \checkmark & 62.10\\
    \hline
    \checkmark & \checkmark & \checkmark & \checkmark & \checkmark & \ding{55} & 62.31 \\
    \checkmark & \checkmark & \checkmark & \checkmark & \ding{55} & - & 61.24 \\
    \checkmark & \checkmark & \checkmark & \ding{55} & - & - & 60.72 \\
    \hline
    \checkmark & \checkmark & \ding{55} & \ding{55} & - & - & 59.76\\
    \checkmark & \ding{55} & \checkmark & \ding{55} & - & - & 60.07 \\
    \checkmark & \ding{55} & \ding{55} & \ding{55} & - & - & 59.21 \\
    \hline
    \multicolumn{6}{l}{\quad Text-only (Diora)} & 58.30 \\    
    \hline
    \end{tabular}
    % }
    \end{center}
    % \vspace{-3mm}
\end{table}
\begin{figure}
    \centering
    \includegraphics[width=0.9\linewidth]{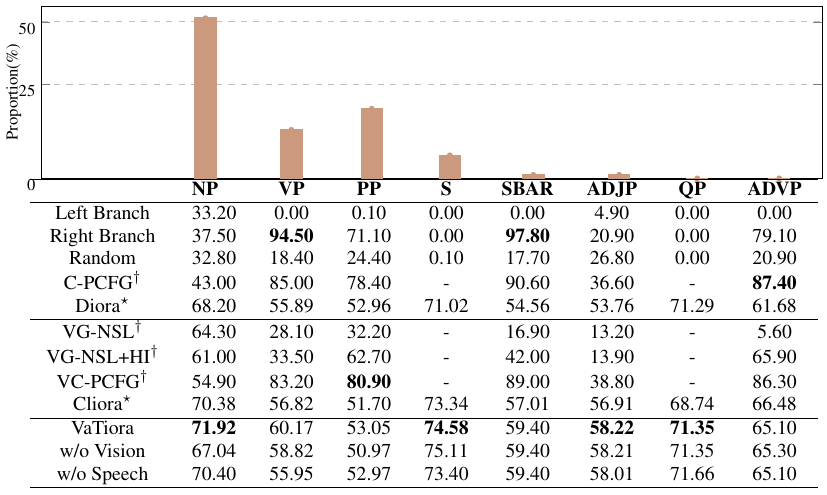}
    \caption{Recall of eight constituent labels on SpokenCOCO test set. The best mean number in each column is in bold. \dag\ indicates results reported by \citet{DBLP:conf/emnlp/ZhaoT20}. $\star$ indicates the results are obtained by running their code on our dataset. The bar chart describes the distribution of each label.}
    \label{fig:enter-label}
\end{figure}
\subsection{Module Ablation}
% \vspace{-1mm}
We display the quantified contributions of each component of VaTiora in Table \ref{tab:ablation}.
The results reveal that speech features and visual features contribute 2.63 and 1.98 corpus-F1, respectively, with speech exhibiting a more significant impact.
More concretely, we conduct ablations on fine-grained speech and vision features individually.
For speech, the pitch and voice activity features collectively contribute 1.32 F1, with pitch playing a more substantial role. 
Concerning vision, using the feature of the entire image instead of mapping object regions to spans results in a performance decay to 61.24, as the entire image compromises the perception of visual hierarchical structures.
Additionally, the region pairs score (i.e. $\bm{M}$) provides valuable information on high-level region composition. 
Moreover, the cross-attention and LSTM modules in VaTiora enhance the system by facilitating multi-modal interaction and capturing pitch patterns.

\subsection{Analysis and Discussions}

We now take one step further, exploring the task and the proposed method from various angles, so as to gain a deeper understanding.

% \vspace{-1mm}
\paratitle{Performance on Span Labels.}
We present a detailed analysis of the performance of eight frequent constituency labels. 
Figure \ref{fig:label} displays a bar chart indicating the proportion of each span label in SpokenCOCO, while the table provides the recall scores for each method. Overall, our model consistently outperforms other diora-based methods across all span types and achieves optimal performance on NP, S, ADJP, and QP constituents. 
The performance of NPs is generally better than other constituents across most methods, primarily due to their dominant proportion (approximately 50\%). 
Vision-based methods (VC-PCFG and Cliora) notably outperform their language-only counterparts (C-PCFG and Dliora) on NPs. 
This can be attributed to the visual information aiding in optimizing the models to focus more on NPs, which are typically noun phrases describing visual scenes.
It is noteworthy that the right branching approach exhibits remarkably high recall on VPs (94.5\%) and SBARs (97.8\%). 
This can be explained by the fact that in a sentence, the predicate and its object phrase, often located on the right side, combine to form a VP constituent. 
Similarly, for SBARs, the subordinate clause is usually positioned after the main clause, which is effectively captured by the right branching strategy.
We also compare the results when removing visual features or speech features, shown in the last two rows in the table.
The performance on NP spans drops a lot without visual features while keep comparable when only removing speech features.
On the contrary, the speech features contribute a lot to the performance on VP spans.
The results explain the different impacts of the two modalities on different grammar components.

% \vspace{-2mm}
\paratitle{Influence of Constituent Length.}
In Figure \ref{fig:constituent length}, we analyze model performance for constituent lengths on SpokenCOCO and SpokenStory, where on SpokenStory we mainly report the results for super long constituents (more than 20 tokens).
Overall, the performance of all models tends to weaken as the sentence length increases.
% all the models become weaker as the sentence length increases.
Diora-based methods exhibit consistent performance across varying sentence lengths, while VG-NSL experiences a significant drop.
PCFG-based methods initially lag behind when the sentence length is short and maintain stable performance for sentence lengths above 5. 
Notably, our VaTiora consistently outperforms other methods across all scenarios.
For long constituents in the SpokenStory, we evaluate the performance of Diora, Cliora, and VaTiora. 
Since the average length of SpokenStory sentences exceeds 20, we can specifically examine these long constituents that are not present in the SpokenCOCO. 
It can be noted that VaTiora still showcases the best performance among them.
% \vspace{-2mm}

\paratitle{Influence of Speech Quality.}
We explore the influence of speech quality by comparing the model on real human voice and mechanically synthetic speech.
% Fortunately, the public dataset 
Specifically, we evaluate the model on SpokenCOCO (with human voice) and SpeechCOCO \footnote{\url{https://zenodo.org/record/4282267}} (with synthetic speech constructed by test-to-speech (TTS) technology).
Note that we specially select SpeechCOCO as the comparison because it adopts an outdated TTS technology, where the synthetic speech has significant difference to the human voice in speech quality.
As shown in Figure \ref{fig:human_speech}, the VaTiora 
demonstrates superior performance in the real human voice.
This is because synthetic speech in SpeechCOCO exhibits a rigid tone while human voice is more natural and contains more reasonable intonation and rhythm owing to human language intuition.
\begin{table}[!pt]
    % \fontsize{8}{10}\selectfont
    \fontsize{10}{12}\selectfont
        \setlength{\tabcolsep}{7.mm}
        \begin{center}
        % \resizebox{1\textwidth}{!}{
        % \vspace{-4mm}
        \captionof{table}{\small
        Corpus-F1 of VaTiora on the SpokenCOCO with different SNR. Smaller SNR means heavier noise.
        }
        % \vspace{2mm}
        \begin{tabular}{ccc}
        \hline
        \bf SNR (dB) & \bf Corpus-F1 (\%)\\
        \hline
        -5 & 55.31 \\
        0 & 58.24 \\
        5 &  61.92 \\
        10 & 62.04 \\

        \hline
        \end{tabular}
    % }
    \label{tab:noise}
    \end{center}
    % \vspace{-3mm}
\end{table}

Further, we explore the model's performance in a noisy environment.
Table \ref{tab:noise} shows the model performance under different signal-to-noise ratios (SNR).
It is evident that light noise shows a relatively minor impact on performance as it causes minimal corruption to the pitch frequency and the voice activity time \cite{DBLP:conf/icassp/LiuW17}.
However, as the noise becomes substantial enough to contaminate the entire waveform, the model's performance experiences a noticeable decline.

\paratitle{Analysis of Grounding.}
We report the phrase localization accuracy to measure the grounding performance of our model.
We use a similarity score to ground each span and image region:
\begin{equation}
    g(i,j,o_m)=\text{Softmax}(\bm{v}_m^\top\bm{h}_{i,j}^{in}),
\end{equation}
where $\bm{v}_m$ is the object feature in Table \ref{tab:feats}, and $\bm{h}_{i,j}^{in}$ is the span representation in Equation \ref{eq:4}.
Then we use $\mathop{Argmax}_m(g(i,j,o_m))$ to find the grounded object.
% We consider two benchmarks COCO and Flickr30K entities \cite{DBLP:journals/ijcv/PlummerWCCHL17}.
Due to the fact that the COCO dataset has not phrase grounding annotations, we manually build test set containing 1,000 samples.
% For the Flickr30K entities, we use the test-to-speech tool to generated 
The results are shown in Table \ref{tab:grnd}, where our model demonstrates good performance and outperforms previous Cliora method.
\begin{table}[t]
    \fontsize{10}{12}\selectfont
    \setlength{\tabcolsep}{2.5mm}
    
    \begin{center}
    % \resizebox{1\textwidth}{!}{
    % \vspace{-6.5mm}
    \caption{
    Phrase grounding results on the built 1,000 COCO test samples.
    % ``$\star$'' means the number is for the whole OpenImage images.
    % SpokenStory dataset and SpokenCOCO test set.
    }
    % \vspace{-2mm}
    \begin{tabular}{lccc}
    \hline
     & \small{\bf Acc(\%)} \\
    \hline
    Cliora & 61.51 \\
    \bf VaTiora & \bf 62.76 \\
    \bottomrule
    \end{tabular}
    % }
    \label{tab:grnd}
    \end{center}
    % \vspace{-5mm}
\end{table}
\begin{figure*}[!pt]
  \centering
  \includegraphics[width=0.99\linewidth]{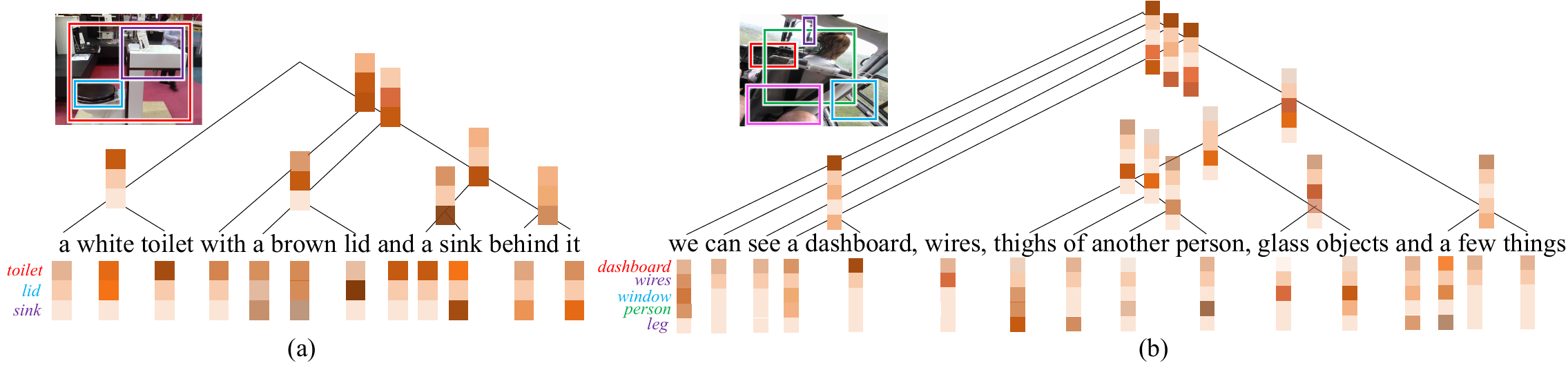}
  % \vspace{-2mm}
  \caption{
  Cases of grounding between spans and regions.
  }
  \label{fig:case}
  % \vspace{-5mm}
\end{figure*}

We also show the visualized analysis for span-region grounding in Figure \ref{fig:case}.
Then we calculate the attention weights ($\text{Softmax}(\bm{v}_m^\top\bm{h}_{i,j}^{in})$) for each span and visualize them in a heat map chart.
Example (a) is chosen from SpokenCOCO, which contains three main objects in a hierarchical structure, i.e., the ``lid'' and the ``sink'' combine into the ``toilet''.
We can see the NP phrases representation could focus on the corresponding object regions, demonstrating a good grounding performance.
The high-level span also has the attention weights to the union regions of its sub-spans.
Example (b) is chosen from SpokenStory, which has longer sentences and more detected objects.
Our model could also allocate a roughly right attention distribution for each span.
For the NPs that have a clear name, such as ``a dashboard'', or ``wires'', the model could learn a sharp attention distribution, while for spans about ``objects'' and ``things'' that represent general objects, it performs poorly to attend to the right targets.
However, this weakness shall be the prevalent problem in unsupervised learning.
Overall, our model could fully make use of the hierarchical structures of images via span-object mapping.

\begin{table}[t]
    % \vspace{-6mm}
    % \fontsize{8}{11.5}\selectfont
    \fontsize{10}{12}\selectfont
    \setlength{\tabcolsep}{6mm}
    
    \begin{center}
    % \resizebox{1\textwidth}{!}{
    % \vspace{-8mm}
    \caption{
    SCF1 results on different word segmentations on SpokenCOCO test set.
    }
    % \vspace{-2mm}
    \begin{tabular}{lccccccc}
    \hline
     \bf Segmentation Methods & \bf tIoU & \bf SCF1\\
    \hline
     VG-HuBert-Golden & 100 & \bf 53.07 \\
     VG-HuBert & 44.02 & 32.50 \\
     VG-W2V2 & 45.56 & 32.42 \\
     HuBERT & 37.72 & 31.59 \\
     W2V2 & 38.56 & 31.77\\
     ResDAVEnet-VQ & 23.78 & 30.02 \\

    \hline
    \end{tabular}
    % }
    \label{tab:word}
    \end{center}
    % \vspace{-3mm}
\end{table}
% \begin{wraptable}{r}{0.45\linewidth}
%     \vspace{-10mm}
%     \fontsize{9}{11.5}\selectfont
%     \setlength{\tabcolsep}{2mm}
    
%     \begin{center}
%     % \resizebox{1\textwidth}{!}{
%     \caption{\small
%     Corpus-F1 results of different word embeddings on SpokenCOCO test set.
%     }
%     % \vspace{-2mm}
%     \begin{tabular}{cccccccc}
%     \hline
%      & \bf Corpus-F1\\
%     \hline
%      Random Initialization & 62.58 \\
%      Elmo & \bf 62.70 \\
%      Glove& 62.49 \\

%     \hline
%     \end{tabular}
%     % }
%     \label{tab:embed}
%     \end{center}
%     \vspace{-3mm}
% \end{wraptable}

\begin{table}[!pt]
% \begin{minipage}[t]{0.45\linewidth}
\fontsize{9}{11.5}\selectfont
    \setlength{\tabcolsep}{2mm}
    
    \begin{center}
    % \resizebox{1\textwidth}{!}{
    \caption{\small
    Corpus-F1 results of different word embeddings on SpokenCOCO test set.
    }
    % \vspace{-2mm}
    \begin{tabular}{cccccccc}
    \hline
     & \bf Corpus-F1\\
    \hline
     Random Initialization & 62.58 \\
     Elmo & \bf 62.70 \\
     Glove& 62.49 \\

    \hline
    \end{tabular}
    % }
    \label{tab:embed}
    \end{center}
\end{table}
% \end{minipage}
% \quad
% \begin{minipage}[t]{0.50\linewidth}
\begin{table}[!pt]
\fontsize{9}{11.5}\selectfont
    \setlength{\tabcolsep}{6mm}
    
    \begin{center}
    % \resizebox{1\textwidth}{!}{
    % \vspace{-8mm}
    \caption{\small
    Corpus-F1 results of different RoI numbers in each image on SpokenCOCO test set.
    }
    % \vspace{-2mm}
    \begin{tabular}{cccccccc}
    \hline
     \bf RoI NUM & \bf Corpus-F1\\
    \hline
     3 & 58.01 \\
     10 & 60.10 \\
     36 & \bf 62.70 \\
     100 & 62.58 \\

    \hline
    \end{tabular}
    % }
    \label{tab:obj}
    \end{center}
% \end{minipage}
\end{table}

\paratitle{Influence of Word Segmentation in Textless Setting}
To explore the influence of word segmentation, we compare the results with predicted word segments in Table \ref{tab:word}.
For evaluation, we adopt the Corpus-F1 to measure the predicted structure and use tIoU to measure the accuracy of word segmentation.
We can see that the model with golden word segmentation could achieve comparable performance (53.07) with text-based GI.
When the accuracy of word segmentation falls, the final F1 score decreases.

\paratitle{Influence of Word Embedding}
We compare the results when using different word embeddings in Table \ref{tab:embed}.
The Elmo outperforms the others, while there is not a very large margin among these methods.
The random embedding could also achieve comparable performance.

\subsection{Influence of Object Detection}
We compare the results of different numbers of RoI in Table \ref{tab:obj}.
We list the results of 3, 10, 36, 100 RoIs per image.
When the RoI number is 36 (which is the default number of Fast-RCNN), the model achieves the best performance.
When the number is too small, i.e., 3 or 10, there is insufficient information for further grammar induction.
When the number is too large, such as 100, there may be more noisy objects with low confidence that decay the performance.

\subsection{Detailed Case Study}
\begin{figure*}
% \vspace{-4.5mm}
\centering
\includegraphics[width=\linewidth]{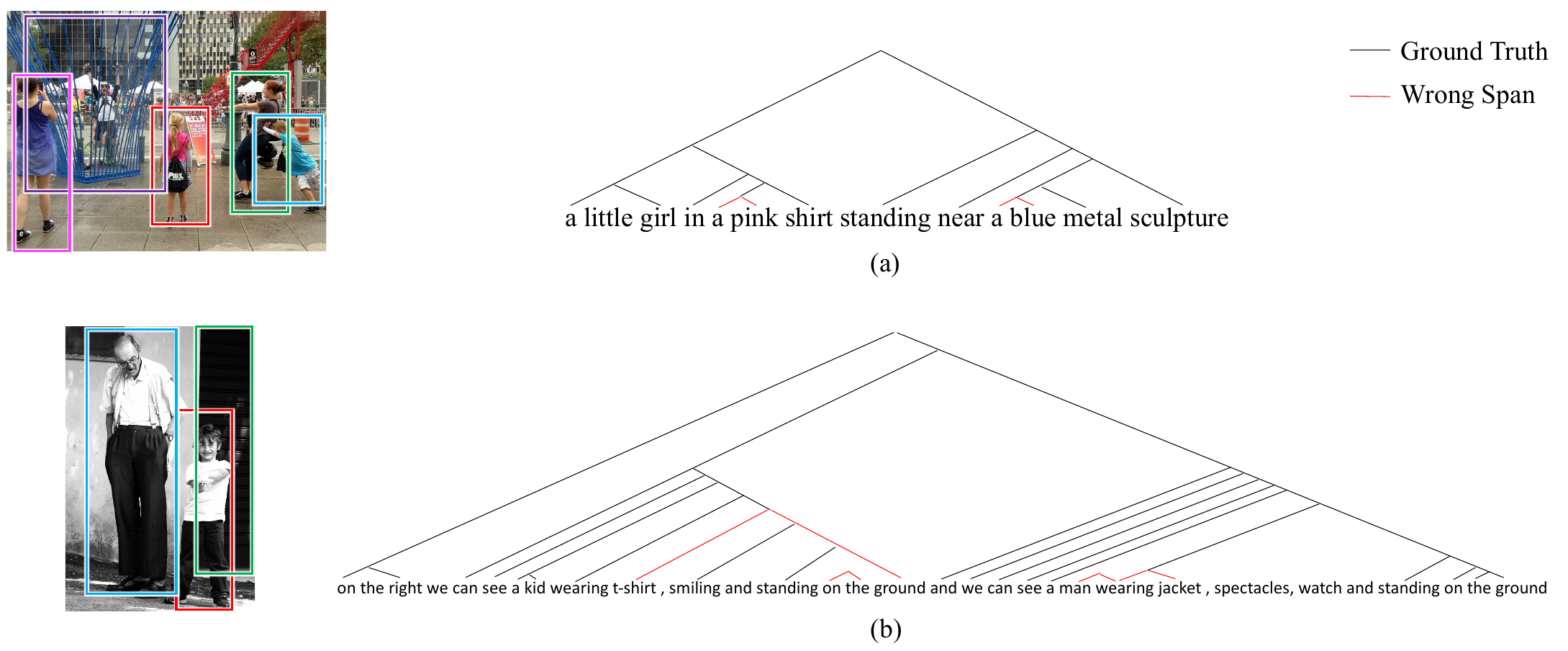}
\caption{
% Inducing the grammar
Case presentation. Example (a) is chosen from SpokenCOCO and example (b) is chosen from SpokenStory.
}
\label{fig:ex_case}
\end{figure*}

\begin{figure*}
% \vspace{-4.5mm}
\centering
\includegraphics[width=\linewidth]{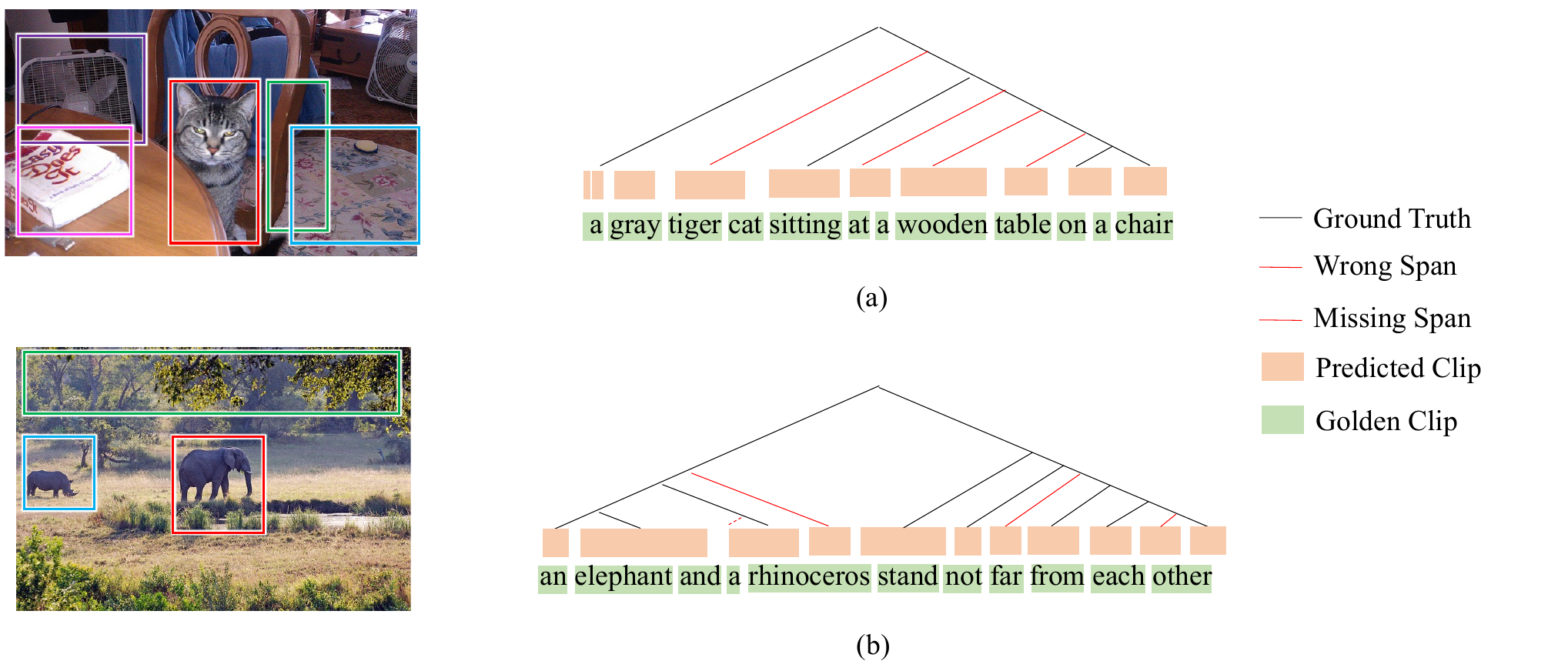}
\caption{
% Inducing the grammar
Case presentation of the textless setting chosen from SpokenCOCO.
}
\label{fig:ex_case2}
% \vspace{-8mm}
\end{figure*}

We visualize the predicted tree structures of VaTiora in Figure \ref{fig:ex_case}, where example (a) is chosen from SpokenCOCO and example (b) is chosen from SpokenStory.
In example (a), VaTiora wrongly predicts the span ``a pink'' and ``a blue'' which should be ``pink shirt'' and ``blue metal''.
For most spans, VaTiora gives the correct prediction.
In example (b), the sentence contains compound sentences and specific symbols.
VaTiora wrongly predicts the span ``t-shirt, smiling and standing on the ground'', which should be ``wearing t-shirt'' and ``smiling and standing on the ground''.
The VaTiora failed to handle the span containing a comma, which seldom appears in the SpokenCOCO training set.
In example (b), we can see the VaTiora could handle the long constituents, such as ``can see ... on the ground''. 
Figure \ref{fig:ex_case2} shows the predicted structures in textless VAT-GI.
We can see the model may predict wrong clip segments, further leading to missing or incorrect spans.
The textless VAT-GI is more challenging due to the lack of text features.
In example (a), the predicted tree degenerates into a right-branching tree.
In example (b), generally, the model predicts a mediocre result, containing some right structures. 


\section{Limitation and Future Work}

In this work, we propose the VAT-GI task that leverages the multi-modal information from image, speech and text to enhance grammar induction.
VAT-GI explores the phylogenetic language acquiring process from the machine learning perspective.
Our proposed VaTiora can properly integrate multi-modal structure features into the grammar parsing, providing a beginning attempt for VAT-GI.
We further discuss the limitations.
First, in VaTiora, the multi-modal feature extraction and grammar parsing is conducted in a pipeline way, which may import noise information due to the limitation of upstream algorithms.
Second, the textless setting of VAT-GI is not fully explored, which may reveal the core mechanism of language acquisition and help enhance the current language models.

As future work of VAT-GI, the following aspects are worth exploring:

% we plan to explore VAT-GI in the following aspects:

{\bf 1) End-to-end multi-modal feature integration.}
Studying the method to extract multi-modal features from inputs in an end-to-end manner.
Currently, the feature extraction in VaTiora tackles three types of inputs, i.e., text, image and speech, respectively and fuse them through a rough vector combination.
In this way, the feature quality depends on the design of upstream algorithms, which have inherent limitations.
A better way to take an end-to-end architecture is to extract the final feature from multiple inputs, where the fusion is processed in an internally implicit way, such that the noise can be avoided.

{\bf 2) Embedding the GI into LLMs for enhanced compositionality.}
Exploring enhancing the large language models (LLMs), and multi-modal LLMs, with the strategy of grammar induction, to help them obtain better performance in multi-modal alignment and strengthen the compositionality ability of language and other modalities.

{\bf 3) Stronger GI methods.}
Proposing stronger methods for grammar induction.
Diora-based methods are designed for text-only setting, which may not adaptive in multi-modal settings.
Also, Diora models the structure through a bottom-up and top-down encoding decoding, which may ignore the latent high-level semantics \cite{DBLP:conf/corr/abs-1904-02142}, thus it is necessary to explore more approaches for grammar induction.

{\bf 4) Exploration on Textless GI.}
In this work, we also introduce the textless setting of VAT-GI which may reveal the core mechanism of language acquisition and help enhance the current language models.
We design our VaTiora system to also support the textless setting.
However, this setting has not been fully explored, and we leave this as a promising potential future work.

% \vspace{-2mm}
\section{Conclusion}

In this paper, we introduce a novel task, visual-audio-text grammar induction (VAT-GI), unsupervisedly inducing the constituent trees from aligned images, text, and speech inputs.
First, we propose a visual-audio-text inside-outside recursive autoencoder (VaTiora) to approach the VAT-GI task, in which the rich modal-specific and complementary features are fully leveraged and effectively integrated. 
To support the textless setting for VAT-GI, we newly devise an aligned span-clip F1 metric (SCF1), which helps conveniently measure the textless constituent structures of the aligned sequences.
Further, we construct a new test set SpokenStory, where the data characterizes richer visual scenes, timbre-rich audio and deeper constituent structures, posing new challenges for VAT-GI.
Experiments on two benchmark datasets indicate that leveraging multiple modality features helps better grammar induction, and also our proposed VaTiora system is more effective in incorporating the various multimodal signals for stronger performance.

\section*{Declaration of Competing Interest}
The authors declare that they have no known competing interests or personal relationships that could have appeared to influence the work reported in this paper.

\section*{Acknowledgments}

This research is supported by National Natural Science Foundation of China (NSFC) under Grant 62336008.

\printcredits

\bibliography{ref}

\end{CJK}
\end{document}